\documentclass[prodmode,acmtkdd]{acmsmall} 

\usepackage{url}
\usepackage{latexsym}
\usepackage{color}
\usepackage[usenames]{xcolor}

\usepackage{cleveref}
\usepackage{amssymb}
\usepackage{amsmath}
\usepackage{graphicx}
\usepackage{rotating}
\usepackage{multirow}
\usepackage{pdfsync}
\usepackage{url}
\usepackage{array}

    \usepackage[applemac]{inputenc}                                          %

  \newif\ifdraft

\newcommand{\ada}{\textsc{AdaBoost.MH}}
\newcommand{\mpb}{\textsc{MP-Boost}}

\newcommand{\hidden}[1]{\vspace{0ex}}

\newcommand{\blue}[1]{\ifdraft{\textcolor{blue}{#1}}\else{\textcolor{black}{#1}}\fi}

\begin{document}

\markboth{G.\ Berardi, A.\ Esuli, and F.\ Sebastiani}{Utility-Theoretic Ranking for
Semi-Automated Text Classification}

\title{Utility-Theoretic Ranking \\ for Semi-Automated Text
Classification} \author{GIACOMO BERARDI, ANDREA ESULI, Italian
National Council of Research \\ FABRIZIO SEBASTIANI, Qatar Computing
Research Institute}

\begin{abstract}
  \emph{Semi-Automated Text Classification} (SATC) may be defined as
  the task of ranking a set $\mathcal{D}$ of automatically labelled
  textual documents in such a way that, if a human annotator validates
  (i.e., inspects and corrects where appropriate) the documents in a
  top-ranked portion of $\mathcal{D}$ with the goal of increasing the
  overall labelling accuracy of $\mathcal{D}$, the expected increase
  is maximized. An obvious SATC strategy is to rank $\mathcal{D}$ so
  that the documents that the classifier has labelled with the lowest
  confidence are top-ranked. In this work we show that this strategy
  is suboptimal. We develop new utility-theoretic ranking methods
  based on the notion of \emph{validation gain}, defined as the
  improvement in classification effectiveness that would derive by
  validating a given automatically labelled document. We also propose
  a new effectiveness measure for SATC-oriented ranking methods, based
  on the expected reduction in classification error brought about by
  partially validating a list generated by a given ranking method.  We
  report the results of experiments showing that, with respect to the
  baseline method above, and according to the proposed measure, our
  utility-theoretic ranking methods can achieve substantially higher
  expected reductions in classification error.
\end{abstract}

\category{Information systems}{Information retrieval}{Retrieval tasks
and goals}[Clustering and Classification] \category{Computing
methodologies}{Machine learning}{Learning paradigms}[Supervised
learning]

\terms{Algorithm, Design, Experimentation, Measurements}




\begin{bottomstuff}

  This paper is a revised and extended version of
  \cite{Berardi:2012fk}.  The order in which the authors are listed is
  purely alphabetical; each author has given an equal contribution to
  this work.

  Authors' address: Giacomo Berardi and Andrea Esuli, Istituto di
  Scienza e Tecnologie dell'Informazione, Consiglio Nazionale delle
  Ricerche, Via Giuseppe Moruzzi 1, 56124 Pisa, Italy. E-mail:
  \emph{firstname.lastname}@isti.cnr.it . Fabrizio Sebastiani, Qatar
  Computing Research Institute, PO Box 5825, Doha, Qatar. E-mail:
  fsebastiani@qf.org.qa . Fabrizio Sebastiani is on leave from the
  Italian National Council of Research.
\end{bottomstuff}

\maketitle


\section{Introduction}
\label{sec:introduction}

\noindent Suppose an organization needs to classify a set
$\mathcal{D}$ of textual documents under classification scheme
$\mathcal{C}$, and suppose that $\mathcal{D}$ is too large to be
classified manually, so that resorting to some form of automated text
classification (TC) is the only viable option. Suppose also that the
organization has strict accuracy standards, so that the level of
effectiveness obtainable via state-of-the-art TC technology (including
any possible improvements obtained via active learning) is not
sufficient.
In this case, the most plausible strategy is to train an automatic
classifier $\hat\Phi$ on the available training data $Tr$, improve it
as much as possible (e.g., via active learning), classify
$\mathcal{D}$ by means of $\hat\Phi$, and then have a human editor
validate (i.e., inspect and correct where appropriate) the results of
the automatic classification. The human annotator will validate only a
subset $\mathcal{D}'\subset\mathcal{D}$, e.g., until she is confident
that the overall level of accuracy of $\mathcal{D}$ is sufficient, or
until she runs out of time.  We call this scenario
\emph{semi-automated text classification} (SATC).

An automatic TC system may support this task by ranking, after the
classification phase has ended and before validation begins, the
classified documents in such a way that, if the human annotator
validates the documents starting from the top of the ranking, the
expected increase in classification effectiveness that derives from
this validation is maximized. This paper is concerned with devising
good ranking strategies for this task.


One obvious strategy \blue{(also used in \cite{Martinez-Alvarez:2012fk})} is to rank the documents in ascending order of
the confidence scores generated by $\hat\Phi$, so that the top-ranked
documents are the ones that $\hat\Phi$ has classified with the lowest
confidence.
The rationale is that an increase in effectiveness can derive only
by validating \emph{misclassified} documents, and that a good ranking
method is simply the one that top-ranks the documents with the highest
probability of misclassification, which (in the absence of other
information) we may take to be the documents which $\hat\Phi$ has
classified with the lowest confidence.

In this work we show that this strategy is, in general,
suboptimal. Simply stated, the reason is that the improvements in
effectiveness that derive from correcting a false positive or a false
negative, respectively, may not be the same, depending on which
evaluation function we take to represent our notion of
``effectiveness''. Additionally, the ratio between these improvements
may vary during the validation process.  In other words, an optimal
ranking strategy must take into account the above improvements and how
these impact on the evaluation function; we will thus look at ranking
methods based on \emph{explicit loss minimization}, i.e., optimized
for the specific effectiveness measures used.
 
The contributions of this paper are the following. First, we develop
new utility-theoretic ranking methods for SATC based on the notion of
\emph{validation gain}, defined as the improvement in effectiveness
that would derive by correcting a given type of mistake (i.e., false
positive or false negative).
Second, we propose a new evaluation measure for SATC based on a
probabilistic user model, and use it to evaluate our experiments on
standard text classification datasets. The results of these
experiments show that, with respect to the confidence-based baseline
method discussed above, our ranking methods are substantially more
effective.

The rest of the paper is organized as follows. Section
\ref{sec:relatedwork} reviews related work, while Section
\ref{sec:definitions} sets the stage by introducing preliminary
definitions and notation. Section \ref{sec:rankingfunction} describes
our base utility-theoretic strategy for ranking the automatically
labelled documents, while in Section \ref{sec:effmeasures} we propose
a novel effectiveness measure for this task based on a probabilistic
user model.  Section \ref{sec:experiments} reports the results of our
experiments in which we test the effectiveness of ranking strategies
by simulating the work of a human annotator that validates
variable-sized portions of the labelled test set.  In Section
\ref{sec:rankingfunctiondynamic} we address a potential problem
deriving from the ``static'' nature of our strategy, by describing a
``dynamic'' (albeit computationally more expensive) version of the
same strategy, and draw an experimental comparison between the two.
In Section \ref{sec:microranking} we acknowledge the existence of two
different ways (``micro'' and ``macro'') of averaging effectiveness
results across classes, and show that the methods we have developed so
far are optimized for macro-averaging; we thus develop and test
methods optimized for micro-averaged effectiveness.  Section
\ref{sec:conclusions} concludes by charting avenues for future
research.
 


\section{Related work}\label{sec:relatedwork}


\noindent Many researchers have tackled the problem of how to improve
on the accuracy delivered by an automatic text classifier when this
accuracy is not up to the standards required by the application (as,
e.g., stipulated in a Service Level Agreement).
%

A standard response to this problem is to ask human annotators to
label additional data that can then be used in retraining a
(hopefully) more accurate classifier. This can be done via the use of
\emph{active learning} techniques (AL -- see e.g.,
\cite{Hoi:2006ef,Tong01}), i.e., via algorithms that rank unlabelled
documents in such a way that the top-ranked ones bring about, once
manually labelled and used for retraining, the highest expected
improvement in classification accuracy. Still, the improvement in
accuracy that can be obtained via active learning is limited: even by
using the best active learning algorithm, accuracy tends to plateau
after a certain number of unlabelled documents have been manually
annotated. When this plateau is reached, annotating more documents
will not improve accuracy any further \cite{Settles:2012uq}.  Similar
considerations apply when active learning is carried out at the term
level, rather than at the document level
\cite{Godbole:2004fk,Raghavan:2006:ALF}.





A related response to the same problem is to use \emph{training data
cleaning} techniques (TDC -- see e.g.,
\cite{Brodley:1999kx,Esuli:2013ko,Fukumoto:2004gf}), i.e., use
algorithms that optimize the human annotator's efforts at correcting
possible labelling mistakes in the training set. TDC algorithms rank
the training documents in such a way that the top-ranked ones bring
about, once their labels are manually checked and then used for
retraining, the highest expected improvement in classification
accuracy. In other words, TDC is to labelled training documents what
AL is to unlabelled ones. Similarly to what happens in active
learning, in many applicative contexts high enough accuracy levels
cannot be attained even at the price of carefully validating the
entire training set for labelling mistakes.

\blue{Yet another response may be the use of some form of \emph{weakly
supervised learning} / \emph{semi-supervised learning}, i.e., of
techniques that allow training a classifier when training data are
few, often leveraging unlabelled data along with the labelled training
data \cite{ChaSchZie06,Zhu:2009fk}.  This solution relies on the fact
that unlabelled data is often available in large quantities, sometimes
even from the same source where the training and test data originate.
Similarly to the cases of AL and TDC, improvements with respect to the
results of the purely supervised setting may be obtained, but these
improvements are going to be limited anyway.}


In conclusion, when the required accuracy standards are high, neither
training data cleaning, nor active learning, nor weakly supervised /
semi-supervised learning, nor a combination of them, may suffice to
reach up to these standards. In this case, after either or all such
techniques have been applied, we can only resort to manual validation
of part of the automatically classified documents by a human
annotator. Supporting this last phase is the goal of semi-automated
text classification.

All the techniques discussed above are different from SATC, since in
SATC we are not concerned with improving the quality of the trained
classifier. We are instead concerned with improving the quality of the
automatically classified \emph{test} set, typically after all attempts
at injecting additional quality in the automatic classifier (and in
the training set) have proved insufficient; in particular, no
retraining / reclassification phase is involved in SATC.


\smallskip

\noindent \textbf{Active learning.} \noindent As remarked above, SATC
certainly bears relations to active learning. In both SATC and in the
\emph{selective sampling} approach to AL (\cite{Lewis94c}; also known
as \emph{pool-based} approach \cite{McCallum98}), the automatically
classified objects are ranked and the human annotator is encouraged to
correct possible misclassifications by working down from the top of
the ranked list. However, as remarked above, the goals of the two
tasks are different. In active learning we are interested in
top-ranking the unlabelled documents that, once manually labelled,
would maximize the information fed back to the learning process, while
in SATC we are interested in top-ranking the unlabelled documents
that, once manually validated, maximize the expected accuracy of the
automatically classified document set. As a result, the optimal
ranking strategies for the two tasks may be different too.

Some approaches to AL take into account the costs of
misclassification, thus attributing different levels of importance to
different types of error.  In \cite{Kapoor:2007} these costs are
embedded into a decision-theoretic framework, which is reminiscent of
our utility-theoretic framework.  A value-of-information criterion is
used in order to select samples which maximize profit, determined by
the total risk of classification and the total cost of labelling.  The
total risk is formulated as a utility function in which the
probability of each classification and the risk associated with it are
taken into account.  The concept of risk is reminiscent of the notion
of ``gain'' defined in our utility function (see Section
\ref{sec:validationgain}), but its purpose is to consider the human
effort needed in correcting a misclassified sample \cite{Vijaya:2009}.
Therefore this decision-theoretic strategy is not aimed to directly
improve classification accuracy, but to minimise the manual work of
the annotator, which is quantified by the risk and the cost of
labelling.

\smallskip

\noindent \textbf{Semi-automated TC.} While AL (and, to a much lesser
degree, TDC) have been investigated extensively in a TC context,
semi-automated TC has been fairly neglected by the research
community. While a number of papers (e.g.,
\cite{Larkey96,ACMCS02,Yang99}) have evoked the existence of this
scenario, we are not aware of \blue{many}  published papers that either discuss
ranking policies for supporting the human annotator's effort, or that
attempt to quantify the effort needed for reaching a desired level of
accuracy. For instance, while discussing a system for the automatic
assignment of ICD9 classes to patients' discharge summaries, Larkey
and Croft {\citeyear{Larkey96} say ``We envision these classifiers
being used in an interactive system which would display the 20 or so
top ranking [classes] and their scores to an expert user. The user
could choose among these candidates (...)'', but do not present
experiments that quantify the accuracy that the validation activity
brings about, or methods aimed at optimizing the cost-effectiveness of
this activity.

\blue{The recent
 \cite{Martinez-Alvarez:2012fk} tackles the related problem of
 deciding when a document is too difficult for automated
 classification, and should thus be routed to a human
 annotator. However, the method presented in the paper is not
 applicable to our case, since (a) it is undefined for documents with
 no predicted labels (a fairly frequent case in multi-label TC), and
 (b) it is undefined when the classification threshold is zero
 (again, a fairly frequent case in modern learning algorithms).}
 
\blue{In a subsequent paper
 \cite{Martinez-Alvarez:2013fk}, the same authors study a family of
 SATC methods that  exploit ``document difficulty'', taking into account
 the confidence scores computed by the base classifiers. They also present
 a comparison between the techniques they propose and that presented
 in an earlier version of the present paper \cite{Berardi:2012fk};
 in this comparison, the former are claimed to outperform the latter
 on the \textsf{Reuters-21578} dataset discussed in Section
 \ref{sec:datasets}. However, this comparison is
 incorrect since the authors compare the results of their ranking
 methods as applied to confidence scores generated by SVMs, 
 with those of the \cite{Berardi:2012fk} ranking method as applied
 to confidences scores generated \emph{by a different learner}. A correct comparison among  ranking  
 methods must instead be
 carried out by providing to all methods the same input, i.e., the
 same confidence scores (whose generation is not part of the method
 itself). 
 The comparison reported in
 \cite{Martinez-Alvarez:2013fk} is incorrect also because it is
 carried out in terms of the $ENER_{\rho}^{\mu}$ measure (see Section \ref{sec:ENER}); instead, as
 stated in \cite{Berardi:2012fk}, the measure according to
 which the method of \cite{Berardi:2012fk} should be
 evaluated is $ENER_{\rho}^{M}$, and not $ENER_{\rho}^{\mu}$, since
 it is $ENER_{\rho}^{M}$ that that method was optimized for. In Section \ref{sec:microranking} we will indeed present SATC methods optimized for $ENER_{\rho}^{\mu}$.}
 
\blue{An application of the method discussed in Section \ref{sec:rankingfunctiondynamic} to performing SATC in a market research context is presented in \cite{Berardi:2014ys}.}

\section{Preliminaries}\label{sec:definitions}

\noindent Given a set of textual documents $\mathcal{D}$ and a
predefined set of classes $\mathcal{C}=\{c_{1}, \ldots, c_{m}\}$,
(multi-class multi-label) TC is usually defined as the task of
estimating an unknown \emph{target function} $\Phi:\mathcal{D}\times
\mathcal{C}\rightarrow\{-1,+1\}$, that describes how documents ought
to be classified, by means of a function $\hat\Phi: \mathcal{D}\times
\mathcal{C}\rightarrow \{-1,+1\}$ called the
\emph{classifier}\footnote{Consistently with most mathematical
literature we use the caret symbol (\^\/\/) to indicate estimation.};
$+1$ and $-1$ represent membership and non-membership of the document
in the class. Here, ``multi-class'' means that there are $m\geq2$
classes, while ``multi-label'' refers to the fact that each document
may belong to zero, one, or several classes at the same
time. Multi-class multi-label TC is usually accomplished by generating
$m$ independent binary classifiers $\hat\Phi_{j}$, one for each $c_{j}
\in \mathcal{C}$, each entrusted with deciding whether a document
belongs or not to a class $c_{j}$.  In this paper we will actually
restrict our attention to classifiers $\hat\Phi_{j}$ that, aside from
taking a binary decision $D_{ij}\in \{-1,+1\}$ on a given document
$d_{i}$, also return a \emph{confidence estimate} $C_{ij}$, i.e., a
numerical value representing the strength of their belief in the fact
that $D_{ij}$ is correct (the higher the value, the higher the
confidence). We formalize this by taking a binary classifier to be a
function $\hat\Phi_{j}: \mathcal{D}\rightarrow \mathbb{R}$ in which
the sign of the returned value $D_{ij}\equiv
sgn(\hat\Phi_{j}(d_{i}))\in \{-1,+1\}$ indicates the binary decision
of the classifier, and the absolute value $C_{ij}\equiv
|\hat\Phi_{j}(d_{i})|$ represents its confidence in the decision.

For the time being we also assume that
\begin{equation}
  \label{eq:F1}
  F_{1}(\hat\Phi_{j}(Te))=\dfrac{2TP_{j}}{2TP_{j}+FP_{j}+FN_{j}}
\end{equation}
(the well-known harmonic mean of precision and recall) is the chosen
evaluation measure for binary classification, where $\hat\Phi_{j}(Te)$
indicates the result of applying $\hat\Phi_{j}$ to the test set $Te$
and $TP_{j}$, $FP_{j}$, $FN_{j}$, $TN_{j}$ indicate the numbers of
true positives, false positives, false negatives, true negatives in
$Te$ for class $c_{j}$.
%
%
%
Note that $F_{1}$ is undefined when $TP_{j}=FP_{j}=FN_{j}=0$; in this
case we take $F_{1}(\hat\Phi_{j}(Te))=1$, since $\hat\Phi_{j}$ has
correctly classified all documents as negative examples. The
assumption that $F_{1}$ is our evaluation measure is not restrictive;
as will be evident later on in the paper, our methods can be
customized to any evaluation function that can be computed from a
contingency table.

As a measure of effectiveness for multi-class multi-label TC, for the
moment being we use \emph{macro-averaged} $F_{1}$ (noted $F_{1}^{M}$),
which is obtained by computing the class-specific $F_{1}$ values and
averaging them across all the $c_{j}\in \mathcal{C}$. An alternative
way of averaging across the classes (\emph{micro-averaged} $F_{1}$)
will be discussed in Section \ref{sec:microranking}.

In this paper the set of unlabelled documents that the classifier must
automatically label (and rank) in the ``operational'' phase will be
represented by the test set $Te$.


\section{A ranking method for SATC based on utility
theory}\label{sec:rankingfunction}

\subsection{Ranking by utility}\label{sec:utility}

\noindent For the time being let us concentrate on the binary case,
i.e., let us assume there is a single class $c_{j}$ that needs to be
separated from its complement $\overline{c}_{j}$. The policy we
propose for ranking the automatically labelled documents in
$\hat\Phi_{j}(Te)$ makes use of \emph{utility theory}, an extension of
probability theory that incorporates the notion of \emph{gain} (or
\emph{loss}) that derives from a given course of action
\cite{Anand:1993fk,von-Neumann:1944uq}.  Utility theory is a general
theory of rational action under uncertainty, and as such is used in
many fields of human activity; for instance, one such field is
betting, since in placing a certain bet we take into account (a) the
probabilities of occurrence that we subjectively attribute to a set of
outcomes (say, to the possible outcomes of a given football game), and
(b) the gains or losses that we obtain, having bet on one of them, if
the various outcomes materialise.

In order to explain our method let us introduce some basics of utility
theory. Given a set $A=\{\alpha_{1}, \alpha_{2}, \ldots\}$ of possible
courses of action and a set $\Omega=\{\omega_{1}, \omega_{2},
\ldots\}$ of mutually disjoint events, the \emph{expected utility}
$U(\alpha_{i}, \Omega)$ that derives from choosing course of action
$\alpha_{i}$ given that any of the events in $\Omega$ may occur, is
defined as
\begin{equation}
  \label{eq:expectedutility}
  U(\alpha_{i}, \Omega)=\sum_{\omega_{k}\in\Omega}P(\omega_{k})G(\alpha_{i},\omega_{k})
\end{equation}
\noindent where $P(\omega_{k})$ is the probability of occurrence of
event $\omega_{k}$ and $G(\alpha_{i},\omega_{k})$ is the \emph{gain}
obtained if $\alpha_{i}$ is chosen and event $\omega_{k}$ occurs. For
instance, $\alpha_{i}$ may be the course of action ``betting on
Arsenal FC's win'' and $\Omega$ may be the set of mutually disjoint
events $\Omega=\{\omega_{1}, \omega_{2}, \omega_{3}\}$, where
$\omega_{1}$=``Arsenal FC wins'', $\omega_{2}$=``Arsenal FC and
Chelsea FC tie'', and $\omega_{3}$=``Chelsea FC wins''; in this case,

\begin{itemize}

\item $P(\omega_{1})$, $P(\omega_{2})$, $P(\omega_{3})$ are the
  probabilities of occurrence that we subjectively attribute to the
  three events $\omega_{1}$, $\omega_{2}$, $\omega_{3}$;
  
\item $G(\alpha_{i}, \omega_{1})$, $G(\alpha_{i}, \omega_{2})$,
  $G(\alpha_{i}, \omega_{3})$ are the economic rewards we obtain if we
  choose course of action $\alpha_{i}$ (i.e., we bet on the win of
  Arsenal FC) and the respective event occurs. Of course, this
  economic reward will be positive if $\omega_{1}$ occurs and negative
  if either $\omega_{2}$ or $\omega_{3}$ occur.

\end{itemize}

\noindent When we face alternative courses of action, acting
rationally means choosing the course of action that maximises our
expected utility. For instance, given the alternative courses of
action $\alpha_{1}$=``betting on Arsenal FC's win'',
$\alpha_{2}$=``betting on Arsenal FC's and Chelsea FC's tie'',
$\alpha_{3}$=``betting on Chelsea FC's win'', we should pick among
$\{\alpha_{1},\alpha_{2},\alpha_{3}\}$ the course of action that
maximises $U(\alpha_{i}, \Omega)$.

How does this translate into a method for ranking automatically
labelled documents? Assume we have a set $D=\{d_{1}, ..., d_{n}\}$ of
such documents that we want to rank, and that $c_{j}$ is the class we
deal with. For instantiating Equation \ref{eq:expectedutility}
concretely we need

\begin{enumerate}

\item \label{item:courses} to decide what our set $A=\{\alpha_{1},
  \alpha_{2}, \ldots\}$ of alternative courses of action is;

\item \label{item:events} to decide what the set $\Omega=\{\omega_{1},
  \omega_{2}, \ldots\}$ of mutually disjoint events is;

\item \label{item:gains} to define the gains $G(\alpha_{i},
  \omega_{k})$;

\item \label{item:probabilities} to specify how we compute the
  probabilities of occurrence $P(\omega_{k})$.

\end{enumerate}

\noindent Let us discuss each of these steps in turn.

Concerning Step \ref{item:courses}, we will take the action of
validating document $d_{i}$ as course of action $\alpha_{i}$. In this
way we will evaluate the expected utility $U_{j}(d_{i},\Omega)$ (i.e.,
the expected increase in the overall classification accuracy of $Te$)
that derives to the classification accuracy of class $c_{j}$ from
validating each document $d_{i}$, and we will be able to rank the
documents by their $U_{j}(d_{i},\Omega)$ value, so as to top-rank the
ones with the highest expected utility.

Concerning Step \ref{item:events}, we have argued in the introduction
that the increase in accuracy that derives from validating a document
depends on whether the document is a true positive, a false positive,
a false negative, or a true negative; as a consequence, we will take
$\Omega=\{tp_{j},fp_{j},fn_{j},tn_{j}\}$, where each of these events
implicitly refers to the document $d_{i}$ under scrutiny (e.g.,
$tp_{j}$ denotes the event ``document $d_{i}$ is a true positive for
class $c_{j}$''). Our utility function has thus the form
\begin{equation}
  \label{eq:ourexpectedutility}
  U_{j}(d_{i}, \Omega)=\sum_{\omega_{k}\in\{tp_{j},fp_{j},fn_{j},tn_{j}\}}P(\omega_{k})G(d_{i},\omega_{k})
\end{equation}
\noindent How to address Step \ref{item:gains} (defining the gains)
will be the subject of Sections \ref{sec:validationgain} and
\ref{sec:smoothing}, while Step \ref{item:probabilities} (computing
the probabilities of occurrence) will be discussed in Section
\ref{sec:estimatingprobabilities}.


\subsection{Validation gains}\label{sec:validationgain}

\noindent We equate $G(d_{i},fp_{j})$ in Equation
\ref{eq:ourexpectedutility} with the average increase in
$F_{1}(\hat\Phi_{j}(Te))$ that would derive by manually validating the
label attributed by $\hat\Phi_{j}$ to a document $d_{i}$ in
$FP_{j}$. We call this the \emph{validation gain} of a document in
$FP_{j}$. Note that validation gains are independent of a particular
document, i.e., $G(d',fp_{j})=G(d'',fp_{j})$ for all $d',d''\in Te$.
Analogous arguments apply to $G(d_{i},tp_{j})$, $G(d_{i},fn_{j})$, and
$G(d_{i},tn_{j})$.

Quite evidently, $G(d_{i},tp_{j})=G(d_{i},tn_{j})=0$, since when the
human annotator validates the label attributed to $d_{i}$ by
$\hat\Phi_{j}$ and finds out it is correct, she will not modify it,
and the value of $F_{1}(\hat\Phi_{j}(Te))$ will thus remain unchanged.

Concerning misclassified documents, it is easy to see that, in
general, $G(d_{i},fp_{j})\not = G(d_{i},fn_{j})$. In fact, if a false
positive is corrected, the increase in $F_{1}$ is the one deriving
from removing a false positive and adding a true negative, i.e.,
\begin{equation}
  \begin{aligned}
    \label{eq:gainFP}
    G(d_{i},fp_{j}) & = \frac{1}{FP_{j}} (F_{1}^{FP}(\hat\Phi_{j}(Te))-F_{1}(\hat\Phi_{j}(Te))) \\
    & = \frac{1}{FP_{j}} (\frac{2TP_{j}}{2TP_{j}+FN_{j}} - \
    \frac{2TP_{j}}{2TP_{j}+FP_{j}+FN_{j}})
  \end{aligned}
\end{equation}
where by $F_{1}^{FP}(\hat\Phi_{j})$ we indicate the value of $F_{1}$
that would derive by correcting all false positives of
$\hat\Phi_{j}(Te)$, i.e., turning all of them into true
negatives. Conversely, if a false negative is corrected, the increase
in $F_{1}$ is the one deriving from removing a false negative and
adding a true positive, i.e.,
\begin{equation}
  \begin{aligned}
    \label{eq:gainFN}
    G(d_{i},fn_{j}) & = \frac{1}{FN_{j}}
    (F_{1}^{FN}(\hat\Phi_{j}(Te))-F_{1}(\hat\Phi_{j}(Te))) \\ & =
    \frac{1}{FN_{j}} (\frac{2(TP_{j}+FN_{j})}{2(TP_{j}+FN_{j})+FP_{j}}
    - \ \frac{2TP_{j}}{2TP_{j}+FP_{j}+FN_{j}})
  \end{aligned}
\end{equation}
where by $F_{1}^{FN}(\hat\Phi_{j})$ we indicate the value of $F_{1}$
that would derive by turning all the false negatives of
$\hat\Phi_{j}(Te)$ into true positives.

\blue{Equation \ref{eq:gainFP} defines the gain deriving from the
correction of a false positive as the \emph{average} across the gains
deriving from the correction of each false positive in the contingency
table (and analogously for Equation \ref{eq:gainFN}). The advantage of
such a definition is that such average gain can be computed once for
all during the entire process. We will see a different definition,
leading to a different SATC method, in Section
\ref{sec:rankingfunctiondynamic}.}

%


\subsection{Smoothing contingency cell estimates}\label{sec:smoothing}

\noindent One problem that needs to be tackled in order to compute
$G(d_{i},fp_{j})$ and $G(d_{i},fn_{j})$ is that the contingency cell
counts $TP_{j}$, $FP_{j}$, $FN_{j}$ are not known (since in
operational settings we do not know which test documents have been
classified correctly and which have been instead misclassified), and
thus need to be estimated\footnote{\label{fn:truenegatives}We will
disregard the estimation of $TN_{j}$ since it is unnecessary for our
purposes, given that $F_{1}(\hat\Phi_{j}(Te))$ does not depend on
$TN_{j}$.}.  In order to estimate them we make the assumption that the
training set and the test set are independent and identically
distributed. We then perform a $k$-fold cross-validation ($k$-FCV) on
the training set: if by $TP_{j}^{Tr}$ we denote the number of true
positives for class $c_{j}$ resulting from the $k$-fold
cross-validation on $Tr$, the maximum-likelihood estimate of $TP_{j}$
is $\hat{TP}_{j}^{ML}=TP_{j}^{Tr}\cdot |Te|/|Tr|$; same for
$\hat{FP}_{j}^{ML}$ and $\hat{FN}_{j}^{ML}$\footnote{\blue{As in many
other contexts, the assumption that the training set and the test set
are independent and identically distributed may not be verified in
practice; if it is not, in our case this leads to imprecise estimates
of the contingency cell counts. While this may be suboptimal, there is
practically nothing that we can do about it, since we do not know the
real values of these counts; in} \blue{other words, $k$-FCV is our
``best possible shot'' at estimating them in the absence of
foreknowledge. As discussed in Section \ref{sec:bounds}, we will
exactly measure \emph{how} suboptimal using $k$-FCV is, by running
experiments in which an oracle feeds our utility-theoretic method with
the true values of the contingency cells.}}.

However, these maximum-likelihood cell count estimates need to be
smoothed, so as to avoid zero counts. In fact, if
$\hat{TP}_{j}^{ML}=0$ it would derive from Equation \ref{eq:gainFP}
that there is nothing to be gained by correcting a false positive,
which is counterintuitive. Similarly, if $\hat{FP}_{j}^{ML}=0$ the
very notion of $F_{1}^{FP}(\hat\Phi_{j})$ would be meaningless, since
it does not make sense to speak of ``removing a false positive'' when
there are no false positives; and the same goes for
$\hat{FN}_{j}^{ML}$.

A second reason why $\hat{TP}_{j}^{ML}$, $\hat{FP}_{j}^{ML}$,
$\hat{FN}_{j}^{ML}$ need to be smoothed is that, when $|Te|/|Tr|<1$,
they may give rise to negative values for $G(d_{i},fp_{j})$ and
$G(d_{i},fn_{j})$, which is counterintuitive. To see this, note that
$\hat{TP}_{j}^{ML}$, $\hat{FP}_{j}^{ML}$, $\hat{FN}_{j}^{ML}$ may not
be integers (which is not bad per se, since the notions of precision,
recall, and their harmonic mean intuitively make sense also when we
allow the contingency cell counts to be nonnegative reals instead of
the usual integers), and may be smaller than 1 (this happens when
$|Te|/|Tr|<1$). This latter fact is problematic, both in theory (since
it is meaningless to speak of, say, removing a false positive from
$Te$ when ``there are less than 1 false positives in $Te$'') and in
practice (since it is easy to verify that negative values for
$G(d_{i},fp_{j})$ and $G(d_{i},fn_{j})$ may derive).

Smoothing has extensively been studied in language modelling for
speech processing \cite{Chen:1996fk} and for ad hoc search in IR
\cite{Zhai:2004fk}. However, the present context is slightly
different, in that we need to smooth contingency tables, and not (as
in the cases above) language models. In particular, while the
$\hat{TP}_{j}^{ML}$, $\hat{FP}_{j}^{ML}$, and $\hat{FN}_{j}^{ML}$ are
the obvious counterparts of the document model resulting from
maximum-likelihood estimation, there is no obvious counterpart to the
``collection model'', thus making the use of, e.g., Jelinek-Mercer
smoothing problematic.
A further difference is that we here require the smoothed counts not
only to be nonzero, but also to be $\geq 1$ (a requirement not to be
found in language modelling).

Smoothing has also been studied specifically for the purpose of
smoothing contingency cell estimates
\cite{Burman:1987fk,Simonoff:1983uq}. However, these methods are
inapplicable to our case, since they were originally conceived for
contingency tables characterized by a small (i.e., $\leq 1$) ratio
between the number of observations (which in our case is $|Te|$) and
the number of cells (which in our case is 4); our case is quite the
opposite. Additionally, these smoothing methods do not operate under
the constraint that the smoothed counts should all be $\geq 1$, which
is a hard constraint for us.

For all these reasons, rather than adopting more sophisticated forms
of smoothing, we adopt simple \emph{additive smoothing} (also known as
\emph{Laplace smoothing}), a special case of Bayesian smoothing using
Dirichlet priors \cite{Zhai:2004fk} which is obtained by adding a
fixed quantity to each of $\hat{TP}_{j}^{ML}$, $\hat{FP}_{j}^{ML}$,
$\hat{FN}_{j}^{ML}$. As a fixed quantity we add 1, since it is the
quantity that all our cell counts need to be greater than or equal to
for Equations \ref{eq:gainFP} and \ref{eq:gainFN} to make sense. We
denote the resulting estimates by $\hat{TP}_{j}^{La}$,
$\hat{FP}_{j}^{La}$, $\hat{FN}_{j}^{La}$.  As it will be clear in
Section \ref{sec:experiments} and following, this simple form of
smoothing proves almost optimal, which seems to indicate that there is
not much to be gained by applying more sophisticated smoothing methods
to our problem context.


Note that we apply smoothing in an ``on demand'' fashion, i.e., we
check if the contingency table needs smoothing at all (i.e., if any of
$\hat{TP}_{j}^{ML}$, $\hat{FP}_{j}^{ML}$, $\hat{FN}_{j}^{ML}$ is $<
1$) and we smooth it only if this is the case.  The reason why we
adopt this ``on-demand'' policy will be especially apparent in Section
\ref{sec:rankingfunctiondynamic}.

\subsection{Turning confidence scores into
probabilities}\label{sec:estimatingprobabilities}

\noindent We derive the probabilities $P(\omega_{k})$ in Equation
\ref{eq:ourexpectedutility} by assuming that the confidence scores
$C_{ij}$ generated by $\hat\Phi_{j}$ can be trusted (i.e., that the
higher $C_{ij}$, the higher the probability that $D_{ij}$ is correct),
and by applying to $C_{ij}$ a \emph{generalized logistic function}
$f(z)=e^{\sigma z}/(e^{\sigma z}+1)$.
This results in
\begin{equation}
  \begin{aligned}
    \label{eq:probfpfn}
    P(fp_{j}|D_{ij}=+1) & = 1-\displaystyle\frac{e^{\sigma
    C_{ij}}}{e^{\sigma C_{ij}}+1}
    \\
    P(fn_{j}|D_{ij}=-1) & = 1-\displaystyle\frac{e^{\sigma
    C_{ij}}}{e^{\sigma C_{ij}}+1}
  \end{aligned}
\end{equation}
\noindent The generalized logistic function (see Figure
\ref{fig:logistic}) has the effect of monotonically converting scores
ranging on $(-\infty,+\infty)$ into real values in the [0.0,1.0] range
(hence the probabilities of Equation \ref{eq:probfpfn} range on
[0.0,0.5]). When $C_{ij}=0$ (this happens when $\hat\Phi_{j}$ has no
confidence at all in its own decision $D_{ij}$), then
\begin{equation}
  \begin{aligned}
    P(tp_{j}|D_{ij}=+1) &= P(fp_{j}|D_{ij}=+1)=0.5 \\
    P(fn_{j}|D_{ij}=-1) &= P(tn_{j}|D_{ij}=-1)=0.5
  \end{aligned}
\end{equation}
\noindent i.e., the probability of correct classification and the
probability of misclassification are identical. Conversely, we have
\begin{equation}
  \begin{aligned}
    \lim_{C_{ij}\rightarrow +\infty}P(fp_{j}|D_{ij}=+1) & =  0 \\
    \lim_{C_{ij}\rightarrow +\infty}P(fn_{j}|D_{ij}=-1) & = 0
  \end{aligned}
\end{equation}
\noindent i.e., when $\hat\Phi_{j}$ has a very high confidence in its
own decision $D_{ij}$, the probability that $D_{ij}$ is wrong is taken
to be close to 0.

\begin{figure}[t]
  \begin{center}
    \scalebox{.70}[.50]{\includegraphics{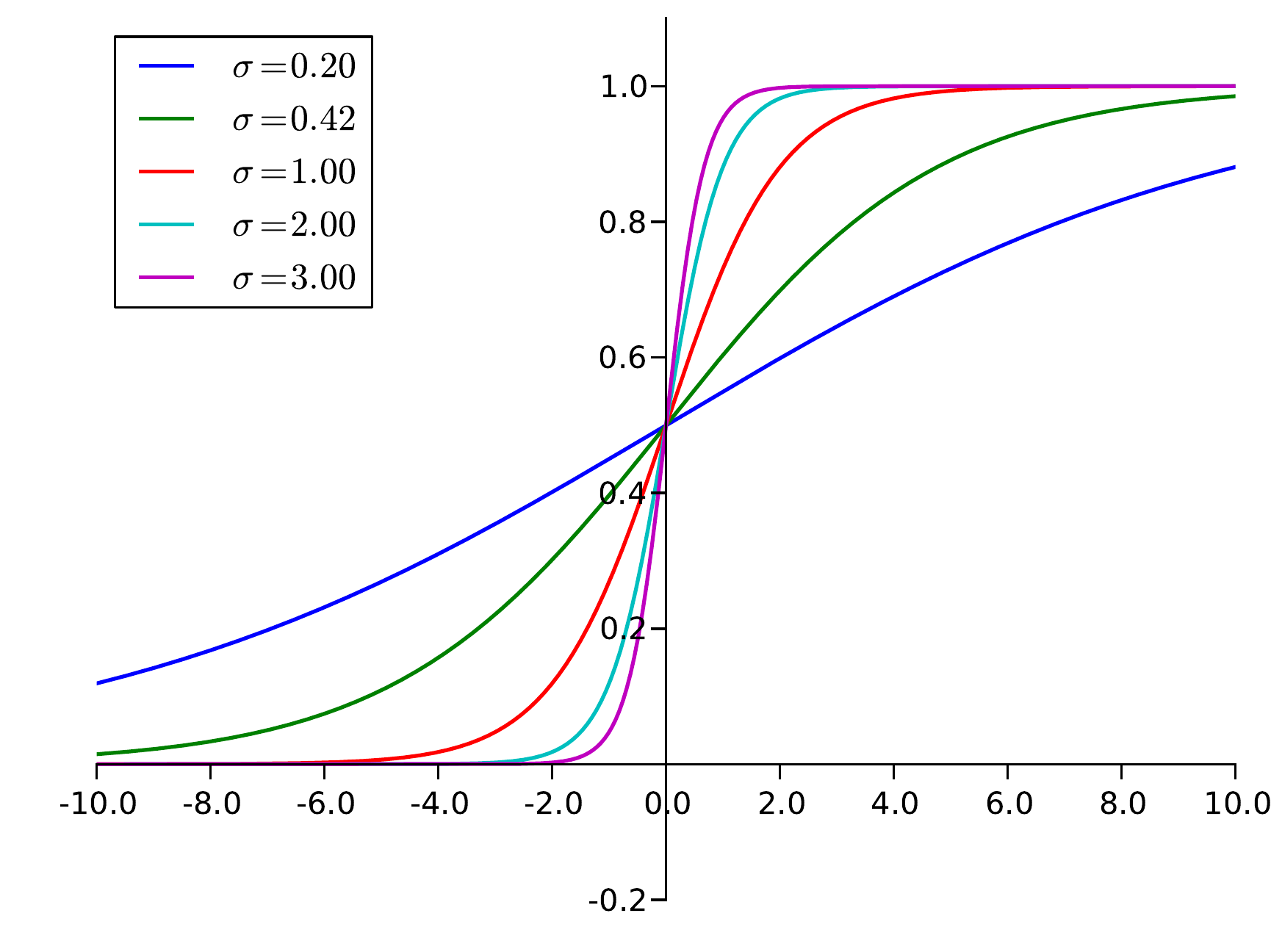}}
  \end{center}
  \caption{\label{fig:logistic}The generalized logistic function.}
\end{figure}

The reason why we use a \emph{generalized} version of the logistic
function instead of \blue{its non-parametric version} (which
corresponds to the case $\sigma=1$) is that using this latter within
Equation \ref{eq:probfpfn} would give rise to a very high number of
zero probabilities of misclassification, since \blue{the
non-parametric logistic function} converts every positive number above
a certain threshold ($\approx$ 36) to a number that standard
implementations round up to 1 even by working in double precision. By
tuning the $\sigma$ parameter (the \emph{growth rate}) we can tune the
speed at which the right-hand side of the sigmoid asymptotically
approaches 1, and we can thus tune how evenly Equation
\ref{eq:probfpfn} distributes the confidence values across the
[0.0,0.5] interval.

The process of optimizing $\sigma$ within Equation \ref{eq:probfpfn}
is usually called \emph{probability calibration}. How we actually
optimize $\sigma$ is discussed in Section \ref{sec:protocol}.


\subsection{Ranking by total utility}\label{sec:rankingbyutility}

\noindent Our function $U_{j}(d_{i},\Omega)$ of Section
\ref{sec:utility} is thus obtained by plugging Equations
\ref{eq:gainFP} and \ref{eq:gainFN} into Equation
\ref{eq:ourexpectedutility}. Therefore, we are now in a position to
compute, given an automatically classified document $d_{i}$ and a
class $c_{j}$, the utility, for the aims of increasing
$F_{1}(\hat\Phi_{j}(Te))$, of manually validating the label $D_{ij}$
attributed to $d_{i}$ by $\hat\Phi_{j}$.

Now, let us recall from Section \ref{sec:definitions} that our goal is
addressing not just the binary, but the multi-class multi-label TC
case, in which binary classification must be accomplished
simultaneously for $|\mathcal{C}|\geq 2$ different classes. It might
seem sensible to propose ranking, for each $c_{j} \in \mathcal{C}$,
all the automatically labelled documents in $Te$ in decreasing order
of their $U_{j}(d_{i},\Omega)$ value.  Unfortunately, this would
generate $|\mathcal{C}|$ different rankings, and in an operational
context it seems implausible to ask a human annotator to scan
$|\mathcal{C}|$ different rankings of the same document set (this
would mean reading the same document $|\mathcal{C}|$ times in order to
validate its labels). As suggested in \cite{Esuli:2009th} for active
learning, it seems instead more plausible to generate a \emph{single}
ranking, according to a score $U(d_{i},\Omega)$ that is a function of
the $|\mathcal{C}|$ different $U_{j}(d_{i},\Omega)$ scores. In such a
way, the human annotator will scan this single ranking from the top,
validating all the $|\mathcal{C}|$ different labels for $d_{i}$ before
moving on to another document. As the criterion for generating the
overall utility score $U(d_{i},\Omega)$ we use \emph{total utility},
corresponding to the simple sum
\begin{equation}
  \label{eq:utilitysum}
  U(d_{i},\Omega)=\sum_{c_{j}\in \mathcal{C}}U_{j}(d_{i},\Omega)
\end{equation}
Our final ranking is thus generated by sorting the test documents in
descending order of their $U(d_{i},\Omega)$ score.

From the standpoint of computational cost, this technique is
$O(|Te|\cdot (|\mathcal{C}|+\log|Te|))$, since the cost of sorting the
test documents by their $U(\cdot,\Omega)$ score is $O(|Te|\log|Te|)$,
and the cost of computing the $U(\cdot,\Omega)$ score for $|Te|$
documents and $|\mathcal{C}|$ classes is $O(|Te|\cdot|\mathcal{C}|)$.


\section{Expected normalized error reduction}\label{sec:effmeasures}

\noindent No measures are known from literature for evaluating the
effectiveness of a SATC-oriented ranking method $\rho$. We here
propose such a measure, which we call \emph{expected normalized error
reduction} (denoted $ENER_{\rho}$). In this section we will introduce
$ENER_{\rho}$ in a stepwise fashion.


\subsection{Error reduction at rank}

\noindent Let us first introduce the notion of \emph{residual error at
rank} $n$ (noted $E_{\rho}(n)$), defined as the error that is still
present in the document set $Te$ after the human annotator has
validated the documents at the first $n$ rank positions in the ranking
generated by $\rho$. The value of $E_{\rho}(0)$ is the initial error
generated by the automated classifier, and the value of
$E_{\rho}(|Te|)$ is 0. We assume our measure of error to range on
[0,1]; if so, $E_{\rho}(n)$ ranges on [0,1] too. We will hereafter
call $n$ the \emph{validation depth} (or \emph{inspection depth}).

We next define \emph{error reduction at rank} $n$ to be
\begin{equation}\label{eq:ER}
  ER_{\rho}(n)=\frac{E_{\rho}(0)-E_{\rho}(n)}{E_{\rho}(0)}
\end{equation} 
i.e., a value in [0,1] that indicates the error reduction obtained by
a human annotator who has validated the documents at the first $n$
rank positions in the ranking generated by $\rho$; 0 stands for no
reduction, 1 stands for total elimination of error.

Example plots of the $ER_{\rho}(n)$ measure are displayed in Figure
\ref{fig:Reuters}, where different curves represent different ranking
methods $\rho', \rho'', ...$, and where, for better convenience, the
$x$ axis indicates the fraction $n/|Te|$ of the test set that has been
validated rather than the number $n$ of validated documents. By
definition all curves start at the origin of the axes (i.e, if the
annotator validates 0 test documents, no error reduction is obtained)
and end at the upper right corner of the graph (i.e., if the annotator
validates all the $|Te|$ test documents, a complete elimination of
error is obtained).  More convex (i.e., higher) curves represent
better strategies, since they indicate that a higher error reduction
is achieved for the same amount of manual validation effort.

\begin{figure}[t]
  \begin{center}
    \resizebox{.49\textwidth}{!}{\includegraphics{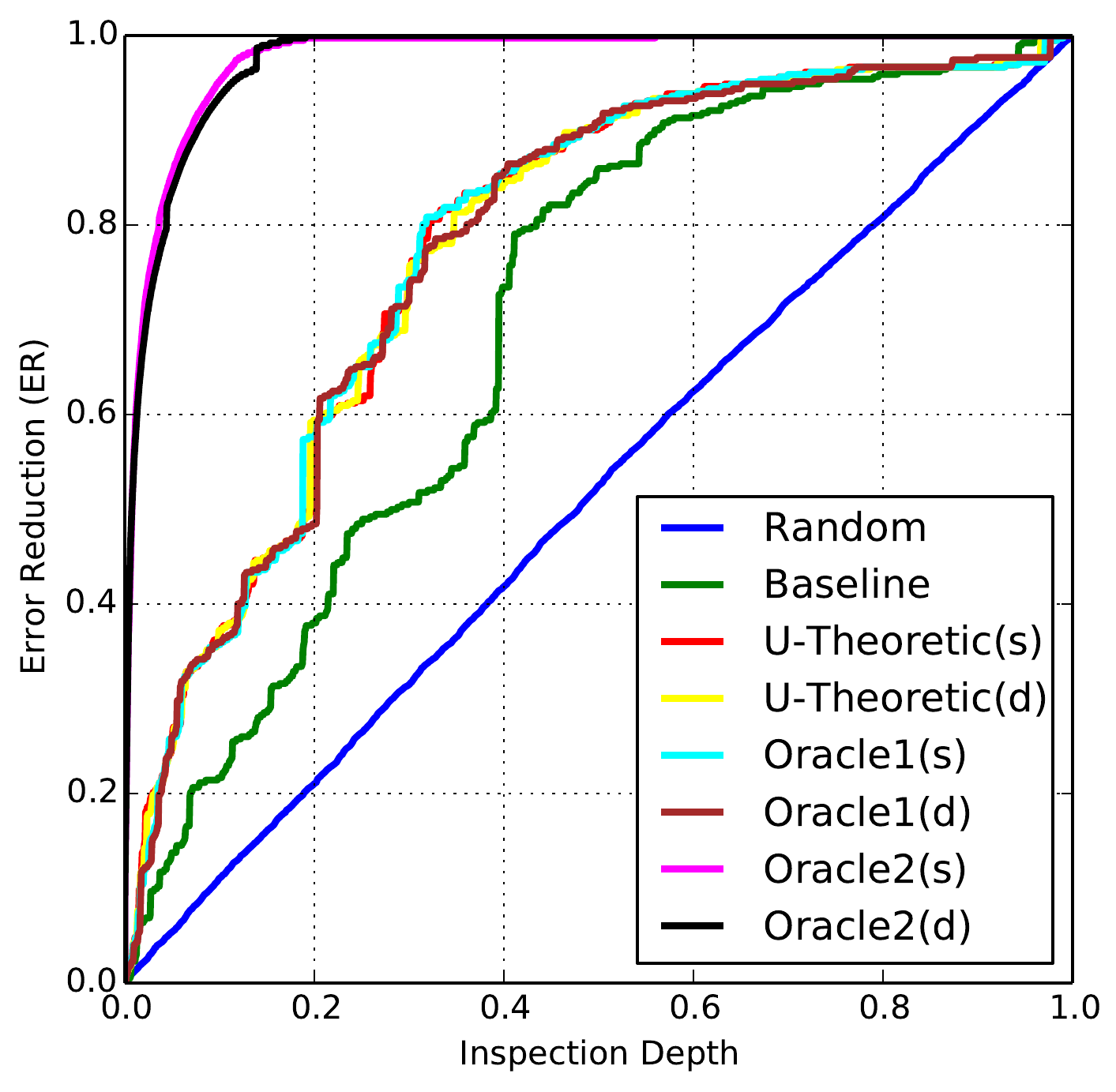}}
    \resizebox{.49\textwidth}{!}{\includegraphics{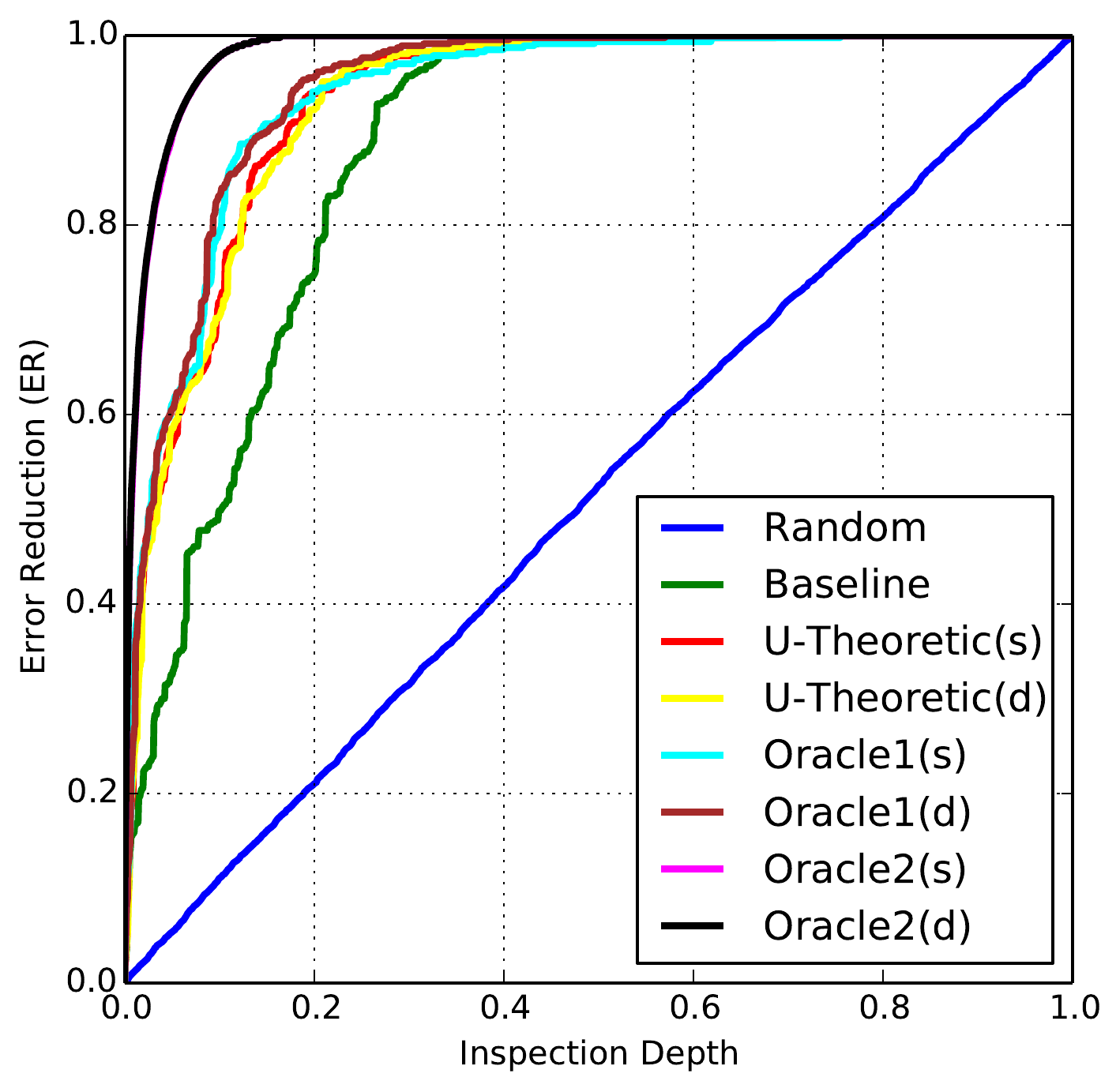}}
  \end{center} \caption{\label{fig:Reuters}Error reduction, measured
  as $ER_{\rho}^{M}$, as a function of validation depth. The dataset
  is \textsc{Reuters-21578}, the learners are \mpb\ (left) and SVMs
  (right). The \textsf{Random} curve indicates the results of our
  estimation of the expected $ER$ of the random ranker via a Monte
  Carlo method with 100 random trials. Higher curves are better.}
\end{figure}

The reason why we focus on error reduction, instead of the
complementary concept of ``increase in accuracy'', is that error
reduction has always the same upper bound (i.e., 100\% reduction),
independently of the initial error.  In contrast, the increase in
accuracy that derives from validating the documents does \emph{not}
always have the same upper bound. For instance, if the initial
accuracy is 0.5, if we assume that accuracy values range on [0,1] then
an increase in accuracy of 100\% is indeed possible, while this
increase is not possible if the initial accuracy is 0.9. This makes
the notion of ``increase in accuracy'' less immediately interpretable,
since different datasets and/or different classifiers give rise to
different initial levels of accuracy. So, using ``error reduction''
instead of ``increase in accuracy'' makes our curves more immediately
interpretable, since error reduction has the same range (i.e., [0,1])
irrespectively of dataset used and/or initial classifier used.

Since (as stated in Section \ref{sec:definitions}) we use $F_{1}$ for
measuring effectiveness, as a measure of classification error we use
$E_{1}\equiv (1-F_{1})$, which indeed (as assumed at the beginning of
this section) ranges on [0,1].  In order to measure the overall
effectiveness of a ranking method across the entire set $\mathcal{C}$
of classes, we compute \emph{macro-averaged} $E_{1}$ (noted
$E_{1}^{M}$), obtained by computing the class-specific $E_{1}$ values
and averaging them across the $c_{j}$'s; from this it derives that
$E_{1}^{M}=1-F_{1}^{M}$.  By $ER_{\rho}^{M}(n)$ we will indicate
macro-averaged $ER_{\rho}(n)$, also obtained by computing the
class-specific $ER_{\rho}(n)$ values and averaging them across the
$c_{j}$'s.





\subsection{Normalized error reduction at rank ...}

\noindent One problem with $ER_{\rho}(n)$, though, is that the
expected $ER_{\rho}(n)$ value of the random ranker is fairly
high\footnote{That the expected $ER_{\rho}(n)$ value of the random
ranker is $\frac{n}{|Te|}$ is something that we have not tried to
formally prove. However, that this holds is supported by intuition
\emph{and} is unequivocally shown by Monte Carlo experiments we have
run on our datasets; see Figures \ref{fig:Reuters} to
\ref{fig:Reuters100} for a graphical representation.}, since it
amounts to $\frac{n}{|Te|}$. The difference between the $ER_{\rho}(n)$
value of a genuinely engineered ranking method $\rho$ and the expected
$ER_{\rho}(n)$ value of the random ranker is particularly small for
high values of $n$, and is null for $n=|Te|$. This means that it makes
sense to factor out the random factor from $ER_{\rho}(n)$. This leads
us to define the \emph{normalized error reduction} of ranking method
$\rho$ as $NER_{\rho}(n)=ER_{\rho}(n)-\frac{n}{|Te|}$, with
macro-averaged $NER_{\rho}(n)$ obtained as usual and denoted, as
usual, by $NER_{\rho}^{M}(n)$.
  

\subsection{... and its expected value}\label{sec:ENER}

\noindent However, $NER_{\rho}(n)$ is still unsatisfactory as a
measure, since it depends on a specific value of $n$ (which is
undesirable, since our human annotator may decide to work down the
ranked list as far as she deems suitable). Following
\cite{Robertson:2008kx} we assume that the human annotator stops
validating the ranked list at exactly rank $n$ with probability
$P_{s}(n)$ (the index $s$ stands for ``stoppage''). We can then define
the \emph{expected normalized error reduction} of ranking method
$\rho$ on a given document set $Te$ as the expected value of
$NER_{\rho}(n)$ according to probability distribution $P_{s}(n)$,
i.e.,
\begin{equation}
  \label{eq:ENER}
  ENER_{\rho}=\sum_{n=1}^{|Te|}P_{s}(n)NER_{\rho}(n)
\end{equation}
%
%
\noindent with macro-averaged $ENER_{\rho}$ indicated, as usual, as
$ENER_{\rho}^{M}$.

Different probability distributions $P_{s}(n)$ can be assumed.
%
%
%
In order to base the definition of such a distribution on a plausible
model of user behaviour, we here make the assumption (along with
\cite{Moffat:2008fk}) that a human annotator, after validating a
document, goes on to validate the next document with probability (or
\emph{persistence} \cite{Moffat:2008fk}) $p$ or stops validating with
probability $(1-p)$, so that
\begin{equation}
  \label{eq:p(1-p)}
  P_{s}(n) = \left \{
    \begin{array}{ll} 
      p^{n-1}(1-p) & \mbox{if $n \in \{1, \ldots, {|Te|-1}$\}} \\ 
      p^{n-1} & \mbox{if $n=|Te|$}
    \end{array} \right . 
\end{equation}
It can be shown that, for a sufficiently large value of $|Te|$,
$\sum_{n=1}^{|Te|}n\cdot P_{s}(n)$ (the expected number of documents
that the human annotator will validate as a function of $p$)
asymptotically tends to $\frac{1}{1-p}$.
The value $\xi=\frac{1}{|Te|(1-p)}$ thus denotes the expected
\emph{fraction} of the test set that the human annotator will validate
as a function of $p$.

Using this distribution in practice entails the need of determining a
realistic value for $p$. A value $p=0$ corresponds to a situation in
which the human annotator only validates the top-ranked document,
while $p=1$ indicates a human annotator who validates each document in
the ranked list.  Unlike in ad hoc search, we think that in a SATC
context it would be unrealistic to take a value for $p$ as given
irrespective of the size of $Te$. In fact, given a desired level of
error reduction, when $|Te|$ is large the human annotators need to be
more persistent (i.e., characterized by higher $p$) than when $|Te|$
is small. Therefore, instead of assuming a predetermined value of $p$
we assume a predetermined value of $\xi$, and derive the value of $p$
from the equation $\xi=\frac{1}{|Te|(1-p)}$. For example, in a certain
application we might assume $\xi=.20$ (i.e., assume that the average
human annotator validates 20\% of the test set). In this case, if
$|Te|=1000$, then $p=1-\frac{1}{.20\cdot 1000}=.9950$, while if
$|Te|=10,000$, then $p=1-\frac{1}{.20\cdot 10000}=.9995$. In the
experiments of Section \ref{sec:experiments} we will test all values
of $p$ corresponding to values of $\xi$ in $\{.05,.10,.20\}$.




Note that the values of $ENER_{\rho}$ are bound above by 1, but a
value of 1 is not attainable. In fact, even the ``perfect ranker''
(i.e., the ranking method that top-ranks all misclassified documents,
noted \emph{Perf}) cannot attain an $ENER_{\rho}$ value of 1, since in
order to achieve total error elimination all the misclassified
documents need to be validated anyway, one by one, which means that
the only condition in which $ENER_{Perf}$ might equal 1 is when there
is just 1 misclassified document. We do not try to normalize
$ENER_{\rho}$ by the value of $ENER_{Perf}$
since $ENER_{Perf}$ cannot be characterized analytically, and depends
on the actual labels in the test set.




\section{Experiments}
\label{sec:experiments}

\noindent We have now fully specified (Section
\ref{sec:rankingfunction}) a method for performing SATC-oriented
ranking and (Section \ref{sec:effmeasures}) a measure for evaluating
the quality of the produced rankings, so we are now in a position to
test the effectiveness of our proposed method. In Sections
\ref{sec:protocol} to \ref{sec:bounds} we will describe our
experimental setting, while in Section \ref{sec:results} we will
report and discuss the actual results of these experiments.


\subsection{Experimental protocol}\label{sec:protocol}

\noindent Let $\Omega$ be a dataset partitioned into a training set
$Tr$ and a test set $Te$. In each experiment reported in this paper we
adopt the following experimental protocol:
\begin{enumerate}
\item For each $c_{j}\in \mathcal{C}$
  \begin{enumerate}
  \item Train classifier $\hat\Phi_{j}$ on $Tr$ and classify $Te$ by
    means of $\hat\Phi_{j}$;
  \item \label{item:compf1} Run $k$-fold cross-validation on $Tr$,
    thereby
    \begin{enumerate}
    \item computing $TP_{j}^{Tr}$, $FP_{j}^{Tr}$, and $FN_{j}^{Tr}$;
    \item \label{step:kFCV} optimizing the $\sigma$ parameter of
      Equation \ref{eq:probfpfn} (see Section \ref{sec:calib} below
      for the actual optimization method used);
    \end{enumerate}
  \end{enumerate}
\item For every ranking policy $\rho$ tested
  \begin{enumerate}
  \item Rank $Te$ according to $\rho$;
  \item Scan the ranked list from the top, correcting possible
    misclassifications and computing the resulting values of
    $ENER_{\rho}^{M}$ for different values of $\xi$.
  \end{enumerate}
\end{enumerate}
\noindent

\noindent For Step \ref{item:compf1} we have used $k=10$\blue{; we
think this value guarantees a good tradeoff between the accuracy of
the parameter estimates (which tends to increase with $k$) and the
cost of computing these estimates (which also increases with $k$)}.

\subsection{Probability calibration}\label{sec:calib}

We optimize the $\sigma$ parameter by picking the value of $\sigma$
that minimizes the average (across the $c_{j}\in \mathcal{C}$)
absolute value of the difference between $Pos^{Tr}_{j}$, the number of
positive training examples of class $c_{j}$, and
$\mathrm{E}[Pos^{Tr}_{j}]$, the \emph{expected} number of such
examples as resulting from the probabilities of membership in $c_{j}$
computed in the $k$-fold cross-validation.  That is, we pool together
all the training documents classified in the $k$-fold cross-validation
phase, and then we pick
\begin{equation}
  \begin{aligned}
    \label{sec:plattcalibration}
    & \arg\min_{\sigma}\displaystyle\frac{1}{|\mathcal{C}|}\displaystyle\sum_{c_{j}\in \mathcal{C}}|Pos^{Tr}_{j}-\mathrm{E}[Pos^{Tr}_{j}]|  = \\
    & \arg\min_{\sigma}\displaystyle\frac{1}{|\mathcal{C}|}\displaystyle\sum_{c_{j}\in \mathcal{C}}|Pos^{Tr}_{j}-\sum_{d_{i}\in Tr}P(c_{j}|d_{i})|  = \\
    &
    \arg\min_{\sigma}\displaystyle\frac{1}{|\mathcal{C}|}\displaystyle\sum_{c_{j}\in
    \mathcal{C}}|Pos^{Tr}_{j}-\sum_{d_{i}\in
    Tr}\frac{e^{\sigma\hat\Phi_{j}(d_{i})}}{e^{\sigma\hat\Phi_{j}(d_{i})}+1}|
  \end{aligned}
\end{equation}
\noindent This method is a much faster calibration method than the
traditional method of picking the value of $\sigma$ that has performed
best in $k$-fold cross-validation\footnote{This method is sometimes
called \emph{Platt calibration} (see e.g.,
\cite{Niculescu-Mizil:2005kx}), due its use in
\cite{Platt:2000fk}. However, the method was in use well before
Platt's article (see e.g., \cite[Section 2.3]{Ittner95}).}.
In fact, unlike the latter, it does not depend on the ranking method
$\rho$. Therefore, this method spares us from the need of ranking the
training set several times, i.e., once for each combination of a
tested value of $\sigma$ and a ranking method $\rho$.




\subsection{Learning algorithms} \label{sec:learners}

\noindent As our first learning algorithm for generating our
classifiers $\hat\Phi_{j}$ we use a boosting-based learner called
\mpb~\cite{Esuli:2006fj}.  Boosting-based methods have shown very good
performance across many learning tasks and, at the same time, have
strong justifications from computational learning theory. \mpb\ is a
variant of \ada\ \cite{Schapire00} optimized for multi-label settings,
which has been shown in~\cite{Esuli:2006fj} to obtain considerable
effectiveness improvements with respect to \ada.
In all our experiments we set the $S$ parameter of \mpb\ (representing
the number of boosting iterations) to 1000.

As the second learning algorithm we use support vector machines
(SVMs). We use the implementation from the freely available LibSvm
library\footnote{\url{http://www.csie.ntu.edu.tw/~cjlin/libsvm/}},
with a linear kernel and parameters at their default values.
 

In all the experiments discussed in this paper stop words have been
removed, punctuation has been removed, all letters have been converted
to lowercase, numbers have been removed, and stemming has been
performed by means of Porter's stemmer. Word stems are thus our
indexing units. Since \mpb\ requires binary input, only their
presence/ absence in the document is recorded, and no weighting is
performed. Documents are instead weighted (by standard
cosine-normalized $tfidf$) for the SVMs experiments.


\subsection{Datasets} \label{sec:datasets}

\noindent Our first dataset is the \textsc{Reuters-21578}
corpus. It consists of a set of 12,902 news stories, partitioned
(according to the standard ``ModApt\'e'' split we have adopted) into a
training set of 9603 documents and a test set of 3299 documents. The
documents are labelled by 118 categories; the average number of
categories per document is 1.08, ranging from a minimum of 0 to a
maximum of 16; the number of positive examples per class ranges from a
minimum of 1 to a maximum of 3964. In our experiments we have
restricted our attention to the 115 categories with at least one
positive training example. This dataset is publicly
available\footnote{\url{http://www.daviddlewis.com/resources/testcollections/~reuters21578/}}
and is probably the most widely used benchmark in text classification
research; this fact allows other researchers to easily replicate the
results of our experiments.

Another dataset we have used is \textsc{OHSUMED} \cite{Hersh94}, a
test collection consisting of a set of 348,566 MEDLINE references
spanning the years from 1987 to 1991.  Each entry consists of summary
information relative to a paper published on one of 270 medical
journals. The available fields are title, abstract, MeSH indexing
terms, author, source, and publication type. Not all the entries
contain abstract and MeSH indexing terms.  In our experiments we have
scrupulously followed the experimental setup presented in
\cite{Lewis96}. In particular, (i) we have used for our experiments
only the 233,445 entries with both abstract and MeSH indexing terms;
(ii) we have used the entries relative to years 1987 to 1990 (183,229
documents) as the training set and those relative to year 1991 (50,216
documents) as the test set; (iii) as the categories on which to
perform our experiments we have used the \emph{main heading} MeSH
index terms assigned to the entries.  Concerning this latter point, we
have restricted our experiments to the 97 MeSH index terms that belong
to the \emph{Heart Disease} (HD) subtree of the MeSH tree, and that
have at least one positive training example. This is the only point in
which we deviate from \cite{Lewis96}, which experiments only on the 77
most frequent MeSH index terms of the HD subtree.

The main characteristics of our datasets, and of three variants
(called \textsc{Reuters-21578/10}, \textsc{Reuters-21578/100}, and
\textsc{OHSUMED-S}) that will be discussed in Section
\ref{sec:results}, are conveniently summarized in Table
\ref{tab:datasets}.

\renewcommand{\tabcolsep}{0.1cm} \renewcommand{\arraystretch}{1.5}
\begin{table}[t]
  \begin{center}
    \tbl{Characteristics of the test collections used. From left
    to right we report the number of test sets $|\mathcal{T}|$ (Column
    2) and, for each test set, the number of training documents $|Tr|$
    (3), the number of test documents $|Te|$ (4), the number of
    classes $|\mathcal{C}|$ (5), and the average number of classes per
    test document $ACD$ (6). Columns 7-10 report the initial error
    (both $E_{1}^{M}$ and $E_{1}^{\mu}$) generated by the \mpb\ and
    SVMs classifiers.}{
    {
    \begin{tabular}{|l||r|r|r|r|r|c|c|c|c|}
      \hline
      \hspace{2.5em}  \multirow{2}{*}{Dataset} & \multirow{2}{*}{$|\mathcal{T}|$} &\multirow{2}{*}{$|Tr|$} \hspace{0.4em} & \multirow{2}{*}{$|Te|$}  \hspace{0.4em} & \multirow{2}{*}{$|\mathcal{C}|$}  \hspace{0.001em} & \multirow{2}{*}{$ACD$}  \hspace{0.001em} &\multicolumn{2}{c|}{$E_{1}^{M}$} & \multicolumn{2}{c|}{$E_{1}^{\mu}$} \\
      \cline{7-10}
      & & & & & & MP-B & SVMs & MP-B & SVMs \\
      \hline\hline
      \textsc{Reuters-21578}     &   1 &   9603 &   3299 & 115 & 1.135 & .392 & .473 & .152 & .140 \\
      \hline
      \textsc{Reuters-21578/10}  &  10 &   9603 &    330 & 115 & 1.135 & .194 & .199 & .151 & .130 \\
      \hline
      \textsc{Reuters-21578/100} & 100 &   9603 &     33 & 115 & 1.135 & .050 & .049 & .149 & .140 \\
      \hline
      \textsc{OHSUMED}           &   1 & 183229 &  50216 &  97 & 0.132 & .553 & .577 & .389 & .324 \\
      \hline
      \textsc{OHSUMED-S}         &   1 &  12358 &   3584 &  97 & 1.851 & .520 & .522 & .286 & .244 \\
      \hline
    \end{tabular}
    }
    }
    \label{tab:datasets}
  \end{center}
\end{table}

%
%
%
%



\subsection{Lower bounds and upper bounds}\label{sec:bounds}

\noindent As the baseline for our experiments we use the
confidence-based strategy discussed in Section \ref{sec:introduction},
which corresponds to using our utility-theoretic method with both
$G(fp)$ and $G(fn)$ set to 1. As discussed in Footnote
\ref{foot:obvious}, while this strategy has not (to the best of our
knowledge) explicitly been proposed before, it seems a reasonable,
common-sense strategy anyway.

While the confidence-based method will act as our lower bound, we have
also run ``oracle-based'' methods aimed at identifying upper bounds
for the effectiveness of our utility-theoretic method, i.e., at
assessing the effectiveness of ``idealized'' (albeit non-realistic)
systems at our task.

The first such method (dubbed \textsf{Oracle1}) works by
``peeking'' at the
actual values of $TP_{j}$, $FP_{j}$, $FN_{j}$ in $Te$, using them in
the computation of $G(d_{i},fp_{j})$ and $G(d_{i},fn_{j})$, and
applying our utility-theoretic method as usual. \textsf{Oracle1} thus
indicates how our method would behave were it able to ``perfectly''
estimate $TP_{j}$, $FP_{j}$, and $FN_{j}$. The difference in
effectiveness between \textsf{Oracle1} and our method will thus be due
to (i) the performance of the method adopted for smoothing contingency
tables, and (ii) possible differences between the distribution of the
documents across the contingency table cells in the training and in
the test set.

In the second such method (\textsf{Oracle2}) we instead
peek \emph{at the true labels} of the documents in $Te$, which means
that we will be able to (a) use the actual values of $TP_{j}$,
$FP_{j}$, $FN_{j}$ in the computation of $G(d_{i},fp_{j})$ and
$G(d_{i},fn_{j})$ (as in \textsf{Oracle1}), \emph{and} (b) replace the
probabilities in Equation \ref{eq:ourexpectedutility} with the true
binary values (i.e., replacing $P(x)$ with 1 if $x$ is true and 0 if
$x$ is false), after which we apply our utility-based ranking method
as usual. The difference in effectiveness between \textsf{Oracle2} and
our method will be due to factors (i) and (ii) already mentioned for
\textsf{Oracle1} \emph{and} to our method's (obvious) inability to
perfectly predict whether a document was classified correctly or not.


\subsection{Results and discussion}\label{sec:results}

%
%
%

\noindent The results of our experiments are given in Table
\ref{tab:allthemacroresults}, where we present the results of running,
for each of two learners (\mpb\ and SVMs) and five datasets
(\textsc{Reuters-21578}, \textsc{OHSUMED}, and three variants of them
-- called \textsc{Reuters-21578/10}, \textsc{Reuters-21578/100},
\textsc{OHSUMED-S} -- that we will introduce in Sections
\ref{sec:smalltestsets}, \ref{sec:tinytestsets},
\ref{sec:largetestsets}), our utility-theoretic method against the
three methods discussed in Section \ref{sec:bounds}. In Table
\ref{tab:allthemacroresults} our method, \textsf{Oracle1} and
\textsf{Oracle2} are actually indicated as \textsf{U-Theoretic(s)},
\textsf{Oracle1(s)} and \textsf{Oracle2(s)}, to distinguish them from
variants (indicated as \textsf{U-Theoretic(d)}, \textsf{Oracle1(d)}
and \textsf{Oracle2(d)}) that will be described in Section
\ref{sec:rankingfunctiondynamic}. Table \ref{tab:allthemacroresults}
presents $ENER_{\rho}^{M}(\xi)$ values for three representative values
of $\xi$, i.e., 0.05, 0.10, and 0.20. 

\begin{table}[h]
  \begin{center}   
  \tbl{\label{tab:allthemacroresults}
  Results of various ranking methods, applied to two learning algorithms and 
  several test collections, in terms of $ENER_{\rho}^{M}(\xi)$, for $\xi\in\{0.05,0.10,0.20\}$. Improvements listed for the various methods are
  relative to the baseline.}{
    \resizebox{\textwidth}{!} {
    \begin{tabular}{|c|l||rr|rr|rr||rr|rr|rr|}
      \hline & & \multicolumn{6}{c||}{\mpb} & \multicolumn{6}{c|}{SVMs} \\
      \hline 
      & &
      \multicolumn{2}{c|}{$\xi=0.05$} &
      \multicolumn{2}{c|}{$\xi=0.10$} &
      \multicolumn{2}{c||}{$\xi=0.20$} &
      \multicolumn{2}{c|}{$\xi=0.05$} &
      \multicolumn{2}{c|}{$\xi=0.10$} &
      \multicolumn{2}{c|}{$\xi=0.20$} \\
      \hline\hline
      \multirow{7}{*}{\begin{sideways}{\textsc{Reuters-21578}}\end{sideways}}
      & Baseline & .071 && .108 && .152 && .262 && .352 && .420 &  \\
      \cline{2-14}
      & U-Theoretic(s) & .163 & (+128\%) & .226 & (+109\%) & .280 & (+84\%) & .442 & (+69\%) & .531 & (+51\%) & .562 & (+34\%)  \\
      \cline{2-14}
      & U-Theoretic(d) & .160 & (+124\%) & .224 & (+107\%) & .279 & (+84\%) & .431 & (+65\%) & .523 & (+49\%) & .557 & (+33\%)  \\
      \cline{2-14}
      & Oracle1(s) & .155 & (+117\%) & .222 & (+106\%) & .280 & (+84\%) & .477 & (+82\%) & .563 & (+60\%) & .586 & (+40\%)  \\
      \cline{2-14}
      & Oracle1(d) & .152 & (+113\%) & .219 & (+103\%) & .275 & (+81\%) & .476 & (+82\%) & .567 & (+61\%) & .592 & (+41\%)  \\
      \cline{2-14}
      & Oracle2(s) & .693 & (+869\%) & .738 & (+583\%) & .707 & (+365\%) & .719 & (+174\%) & .760 & (+116\%) & .723 & (+72\%)  \\
      \cline{2-14}
      & Oracle2(d) & .677 & (+847\%) & .725 & (+571\%) & .699 & (+360\%) & .723 & (+176\%) & .763 & (+117\%) & .724 & (+72\%) \\
      \hline\hline
      \multirow{7}{*}{\begin{sideways}{\textsc{Reuters-21578/10}}\end{sideways}}
      & Baseline & .063 && .097 && .135 && .243 && .322 && .383 &  \\
      \cline{2-14}
      & U-Theoretic(s) & .145 & (+131\%) & .203 & (+110\%) & .245 & (+81\%) & .330 & (+36\%) & .415 & (+29\%) & .465 & (+21\%)  \\
      \cline{2-14}
      & U-Theoretic(d) & .139 & (+121\%) & .198 & (+105\%) & .239 & (+77\%) & .335 & (+38\%) & .420 & (+30\%) & .470 & (+23\%)  \\
      \cline{2-14}
      & Oracle1(s) & .159 & (+153\%) & .205 & (+112\%) & .243 & (+80\%) & .392 & (+61\%) & .482 & (+50\%) & .522 & (+36\%)  \\
      \cline{2-14}
      & Oracle1(d) & .158 & (+152\%) & .212 & (+119\%) & .255 & (+89\%) & .394 & (+62\%) & .488 & (+52\%) & .531 & (+39\%)  \\
      \cline{2-14}
      & Oracle2(s) & .555 & (+784\%) & .643 & (+566\%) & .648 & (+380\%) & .596 & (+145\%) & .676 & (+110\%) & .672 & (+75\%)  \\
      \cline{2-14}
      & Oracle2(d) & .558 & (+789\%) & .648 & (+571\%) & .654 & (+384\%) & .599 & (+147\%) & .679 & (+111\%) & .675 & (+76\%) \\
      \hline\hline
      \multirow{7}{*}{\begin{sideways}{\textsc{Reuters-21578/100}}\end{sideways}}
      & Baseline & .069 && .121 && .164 && .226 && .302 && .364 &  \\
      \cline{2-14}
      & U-Theoretic(s) & .118 & (+71\%) & .172 & (+42\%) & .215 & (+31\%) & .291 & (+29\%) & .365 & (+21\%) & .416 & (+14\%)  \\
      \cline{2-14}
      & U-Theoretic(d) & .119 & (+72\%) & .176 & (+45\%) & .217 & (+32\%) & .289 & (+28\%) & .367 & (+22\%) & .419 & (+15\%)  \\
      \cline{2-14}
      & Oracle1(s) & .192 & (+178\%) & .247 & (+104\%) & .281 & (+71\%) & .318 & (+41\%) & .422 & (+40\%) & .479 & (+32\%)  \\
      \cline{2-14}
      & Oracle1(d) & .197 & (+185\%) & .266 & (+120\%) & .318 & (+94\%) & .318 & (+41\%) & .427 & (+41\%) & .489 & (+34\%)  \\
      \cline{2-14}
      & Oracle2(s) & .429 & (+521\%) & .537 & (+344\%) & .575 & (+251\%) & .458 & (+103\%) & .568 & (+88\%) & .600 & (+65\%)  \\
      \cline{2-14}
      & Oracle2(d) & .429 & (+521\%) & .537 & (+344\%) & .576 & (+251\%) & .458 & (+103\%) & .569 & (+88\%) & .601 & (+65\%)  \\         
      \hline\hline
      \multirow{7}{*}{\begin{sideways}{\textsc{OHSUMED}}\end{sideways}}
      & Baseline & .385 && .479 && .512 && .526 && .630 && .644 &  \\
      \cline{2-14}
      & U-Theoretic(s) & .442 & (+15\%) & .529 & (+10\%) & .549 & (+7\%) & .623 & (+18\%) & .685 & (+9\%) & .666 & (+3\%)  \\
      \cline{2-14}
      & U-Theoretic(d) & .443 & (+15\%) & .531 & (+11\%) & .550 & (+7\%) & .618 & (+17\%) & .676 & (+7\%) & .655 & (+2\%)  \\
      \cline{2-14}
      & Oracle1(s) & .445 & (+16\%) & .530 & (+11\%) & .549 & (+7\%) & .639 & (+21\%) & .687 & (+9\%) & .657 & (+2\%)  \\
      \cline{2-14}
      & Oracle1(d) & .449 & (+17\%) & .532 & (+11\%) & .550 & (+7\%) & .617 & (+17\%) & .659 & (+5\%) & .636 & (-1\%)  \\
      \cline{2-14}
      & Oracle2(s) & .838 & (+118\%) & .839 & (+75\%) & .769 & (+50\%) & .864 & (+64\%) & .854 & (+36\%) & .778 & (+21\%)  \\
      \cline{2-14}
      & Oracle2(d) & .758 & (+97\%) & .762 & (+59\%) & .700 & (+37\%) & .795 & (+51\%) & .787 & (+25\%) & .721 & (+12\%)  \\
      \hline\hline
      \multirow{7}{*}{\begin{sideways}{\textsc{OHSUMED-S}}\end{sideways}}
      & Baseline & .021 && .025 && .026 && .075 && .124 && .164 &  \\
      \cline{2-14}
      & U-Theoretic(s) & .087 & (+323\%) & .118 & (+374\%) & .132 & (+402\%) & .212 & (+184\%) & .282 & (+127\%) & .323 & (+97\%)  \\
      \cline{2-14}
      & U-Theoretic(d) & .088 & (+329\%) & .118 & (+374\%) & .132 & (+402\%) & .210 & (+182\%) & .280 & (+126\%) & .321 & (+96\%)  \\
      \cline{2-14}
      & Oracle1(s) & .091 & (+343\%) & .117 & (+370\%) & .125 & (+375\%) & .272 & (+265\%) & .334 & (+169\%) & .352 & (+115\%)  \\
      \cline{2-14}
      & Oracle1(d) & .094 & (+358\%) & .119 & (+378\%) & .128 & (+387\%) & .301 & (+303\%) & .363 & (+193\%) & .380 & (+132\%)  \\
      \cline{2-14}
      & Oracle2(s) & .481 & (+2246\%) & .554 & (+2125\%) & .572 & (+2075\%) & .511 & (+585\%) & .589 & (+375\%) & .603 & (+268\%)  \\
      \cline{2-14}
      & Oracle2(d) & .450 & (+2095\%) & .498 & (+1900\%) & .496 & (+1786\%) & .487 & (+553\%) & .540 & (+335\%) & .536 & (+227\%)  \\

      \hline\hline
    \end{tabular}
    }
    }
  \end{center}
\end{table}

\blue{For each of two learners and five datasets, and for each pairwise combination of all the methods discussed (including those we will discuss in Section \ref{sec:rankingfunctiondynamic}), we have run a paired t-test with $ENER_{\rho}^{M}$(0.10) as the evaluation measure and $0.05$ as the significance level, in order to determine whether the difference in performance between the two methods is statistically significant. The results of such tests are reported in Table \ref{tab:stat}.} 

\begin{table}[t]
  \begin{center} 
  \tbl{\label{tab:stat}\blue{Statistical significance results obtained for the two learners (MP-Boost and SVMs) via a paired t-test with $ENER_{\rho}^{M}$(0.10) as the evaluation measure and  $0.05$ as the significance level. ``Y'' means that there is a statistically significant difference between the two methods, while ``N'' means there is not; each 5-tuple of Y's and N's indicates this for the five datasets studied in this paper (\textsc{Reuters-21578},
  \textsc{Reuters-21578/10}, \textsc{Reuters-21578/100},
  \textsc{OHSUMED}, \textsc{OHSUMED-S}, in this order).}}{
    \begin{tabular}{|c|c||c|c|c|c|c|c|c|}
      \hline 
      \multicolumn{2}{|c||}{\mbox}  & \begin{sideways}Baseline\end{sideways} & \begin{sideways}U-Theoretic(s)\end{sideways} & \begin{sideways}U-Theoretic(d)\end{sideways} & \begin{sideways}Oracle1(s)\end{sideways} & \begin{sideways}Oracle1(d)\end{sideways} & \begin{sideways}Oracle2(s)\end{sideways} & \begin{sideways}Oracle2(d)\end{sideways} \\
      \hline\hline
      \multirow{7}{*}{\begin{sideways}MP-Boost\end{sideways}}
      & Baseline & $----$ & YYYYY & YYYYY & YYYYY & YYYYY & YYYYY & YYYYY         \\ \cline{2-9}
      & U-Theoretic(s)  & YYYYY & $----$ & NYNNN & NNYNN & YNYNN & YYYYY & YYYYY \\ \cline{2-9}
      & U-Theoretic(d)  & YYYYY & NYNNN & $----$  & NNYNN & NNYNN & YYYYY & YYYYY \\ \cline{2-9}
      & Oracle1(s) & YYYYY &  NNYNN & NNYNN & $----$ & NNYNN & YYYYY & YYYYY     \\ \cline{2-9}
      & Oracle1(d) & YYYYY &  YNYNN & NNYNN & NNYNN & $----$ & YYYYY & YYYYY     \\ \cline{2-9}
      & Oracle2(s) & YYYYY & YYYYY  & YYYYY  & YYYYY & YYYYY & $----$ & NNNYN     \\ \cline{2-9}
      & Oracle2(d) & YYYYY & YYYYY  & YYYYY & YYYYY & YYYYY & NNNYN & $----$      \\ \cline{2-9}
     \hline\hline
      \multirow{7}{*}{\begin{sideways}SVMs\end{sideways}}
      & Baseline & $----$ & YYYYY & YYYYY & YYYYY & YYYYY & YYYYY & YYYYY         \\ \cline{2-9}
      & U-Theoretic(s) & YYYYY & $----$  & YNNYN & YYYNY & YYYNY & YYYYY & YYYYY \\ \cline{2-9}
      & U-Theoretic(d) & YYYYY & YNNYN & $----$  & YYYNY & YYYNY & YYYYY & YYYYY \\ \cline{2-9}
      & Oracle1(s) & YYYYY & YYYNY & YYYNY & $----$  & NNYNY & YYYYY & YYYYY     \\ \cline{2-9}
      & Oracle1(d) & YYYYY & YYYNY & YYYNY & NNYNY & $----$  & YYYYY & YYYYY     \\ \cline{2-9}
      & Oracle2(s) & YYYYY & YYYYY & YYYYY & YYYYY & YYYYY & $----$  & YYNNN     \\ \cline{2-9}
      & Oracle2(d) & YYYYY & YYYYY  & YYYYY & YYYYY & YYYYY & YYNNN &  $----$     \\ \cline{2-9}
      \hline
    \end{tabular}
    }
  \end{center}
\end{table}


\subsubsection{Mid-sized test sets}\label{sec:midsizedtestsets}

\noindent Figure \ref{fig:Reuters} plots the results, in terms of
$ER_{\rho}^{M}(n)$, of our experiments with the \mpb\ and SVM learners
on the \textsc{Reuters-21578} dataset. The results of these
experiments in terms of $ENER_{\rho}^{M}$ as a function of the chosen
value of $\xi$ are instead reported in Table
\ref{tab:allthemacroresults}.
The optimal value of $\sigma$ returned by the $k$-fold
cross-validation phase is $.554$ for \mpb\ and $7.096$ for SVMs; these
values, sharply different from 1 and from each other, clearly show the
advantage of converting confidence scores into probabilities via a
\emph{generalized} logistic function.

The first insight we can draw from these results is that our
\textsf{U-Theoretic(s)} method outperforms \textsf{Baseline} in a very
substantial way \blue{(the paired t-test -- see Table \ref{tab:stat} -- indicates that this difference is  statistically significant)}. This can be appreciated both from the plots of
Figures \ref{fig:Reuters}, in which the red curve (corresponding to
\textsf{U-Theoretic(s)}) is markedly higher than the green curve
(corresponding to \textsf{Baseline}), and from Table
\ref{tab:allthemacroresults}.  In this latter, for $\xi=.10$
(corresponding to $p=.996$) our method obtains relative improvements
over \textsf{Baseline} of +109\% (\mpb) and +51\% (SVMs); for
$\xi=.20$ the improvements, while not as high as for $\xi=.10$, are
still sizeable (+84\% for \mpb\ and +34\% for SVMs), while for
$\xi=.05$ the improvements are even higher than for $\xi=.10$ (+128\%
for \mpb\ and +69\% for SVMs).

A second insight is that, surprisingly, our method hardly differs in
terms of performance from \textsf{Oracle1(s)}. The two curves can be
barely distinguished in Figure \ref{fig:Reuters}, and in terms of
$ENER_{\rho}^{M}$ \textsf{Oracle1(s)} is even slightly outperformed,
in the \mpb\ experiments, by \textsf{U-Theoretic(s)} (e.g., .226 vs.\
.222 for $\xi=.10$); \blue{the paired t-test (see Table \ref{tab:stat}) indicates that the difference between the two methods is not statistically significant.} This shows that (at least judging from these
experiments) Laplace smoothing is nearly optimal, and there is likely
not much we can gain from applying alternative, more sophisticated
smoothing methods. This is sharply different from what happens in
language modelling, where Laplace smoothing has been shown to be an
underperformer \cite{Gale:1994fk}. The fact that with \mpb\ our method
slightly (and strangely) outperforms \textsf{Oracle1(s)} is probably
due to accidental, ``serendipitous'' interactions between the
probability estimation component (Equation \ref{eq:probfpfn}) and the
contingency cell estimation component of Section \ref{sec:smoothing}; \blue{in fact, the paired t-test indicates (see Table \ref{tab:stat}) that this difference is not statistically significant.}

A third interesting fact is that error reduction is markedly better in
the SVM experiments than in the \mpb\ experiments. This is evident
from the fact that the Figure \ref{fig:Reuters} curves for SVMs are
much more convex (i.e., are higher) and are closer to the optimum
(i.e., closer to the \textsf{Oracle2(s)} curve) than the corresponding
Figure \ref{fig:Reuters} curves for \mpb. This fact is also evident
from the numerical results reported in Table
\ref{tab:allthemacroresults} where, with \textsf{U-Theoretic(s)}, SVMs
obtain $ENER_{\rho}^{M}(.10)=.531$, which is +134\% better than the
$ENER_{\rho}^{M}(.10)=.226$ result obtained by \mpb\ (similar
improvements can be observed for the other methods and for the other
values of $\xi$). This provides a striking contrast with the
\emph{classification accuracy} results reported in Figure
\ref{tab:datasets} where, on the same dataset, \mpb\
($E_{1}^{M}=.392$) substantially outperformed SVMs
($E_{1}^{M}=.473$). It is easy to conjecture that, even if \mpb\ yields
higher classification accuracy, it generates less reliable
(calibrated) confidence scores, i.e., it generates confidence scores
that correlate with the ground truth worse than the SVM-generated
scores.

The rates of improvement of \textsf{U-Theoretic(s)} over the baseline
are instead much higher for \mpb\ than for SVMs (e.g., for $\xi=.10$
these are +109\% and +51\%, respectively). (The same goes for the
improvements of \textsf{Oracle1(s)} over the baseline.) This is likely
due to the fact that, as observed above, the absolute values of
$ENER_{\rho}^{M}(\xi)$ obtained by the baseline are much higher for
SVMs than for \mpb\ for all methods, so the margins of improvement
with respect to the baseline are smaller for SVMs than for \mpb.

\subsubsection{Small test sets}\label{sec:smalltestsets}

\noindent We have also run a batch of experiments aimed at assessing
how the methods fare when ranking test sets much smaller than
\textsc{Reuters-21578}. This may be \emph{more} challenging than
ranking larger sets since, when the test set is small, Laplace
smoothing (i) can seriously perturb the relative proportions among the
cell counts, which can generate poor estimates of $G(d_{i},fp_{j})$
and $G(d_{i},fn_{j})$, and (ii) is performed for more classes, since
(as discussed at the end of Section \ref{sec:smoothing}) we smooth
``on demand'' only, and since the likelihood that $\hat{TP}_{j}^{ML}$,
$\hat{FP}_{j}^{ML}$, $\hat{FN}_{j}^{ML}$ are smaller than 1 is higher
with small test sets. This is also a realistic setting since, if a set
of unlabelled documents is small, it is likely that validating a
portion of it that can lead to sizeable enough effectiveness
improvements is feasible from an economic point of view.

Rather than choosing a completely different dataset, we generate 10
new test sets by randomly splitting the \textsc{Reuters-21578} test
set in 10 equally-sized parts (about 330 documents each). In our
experiments we run each ranking method on each such part individually
and average the results across the 10 parts. We call this experimental
scenario \textsc{Reuters-21578/10}. This allows us to study the
effects of test set size on our methods in a more controlled way than
if we had picked a completely different dataset, since test set size
is the only difference with respect to the previous
\textsc{Reuters-21578} experiments.

\begin{figure}[t]
  \begin{center}
    \resizebox{.49\textwidth}{!}{\includegraphics{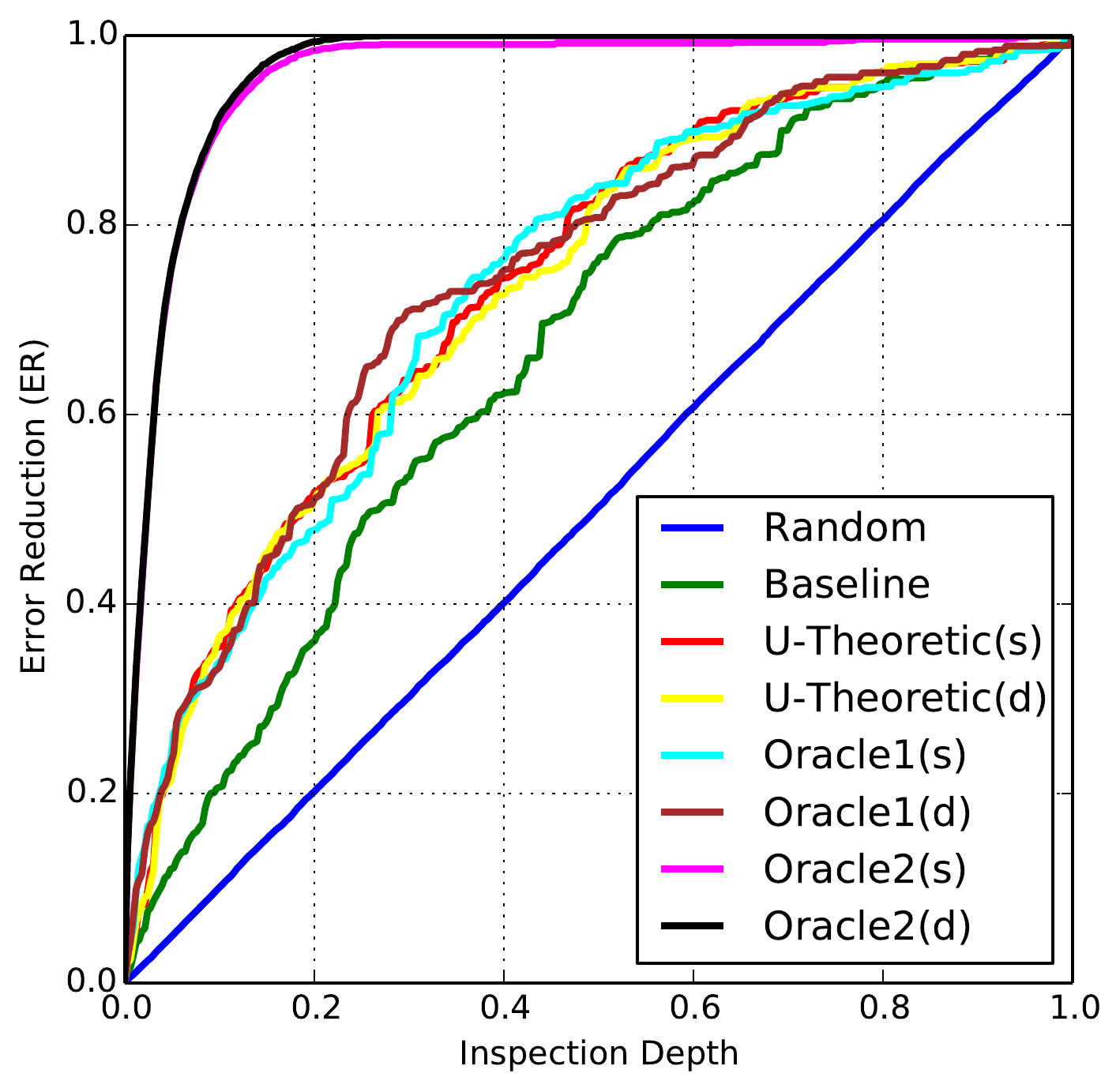}}
    \resizebox{.49\textwidth}{!}{\includegraphics{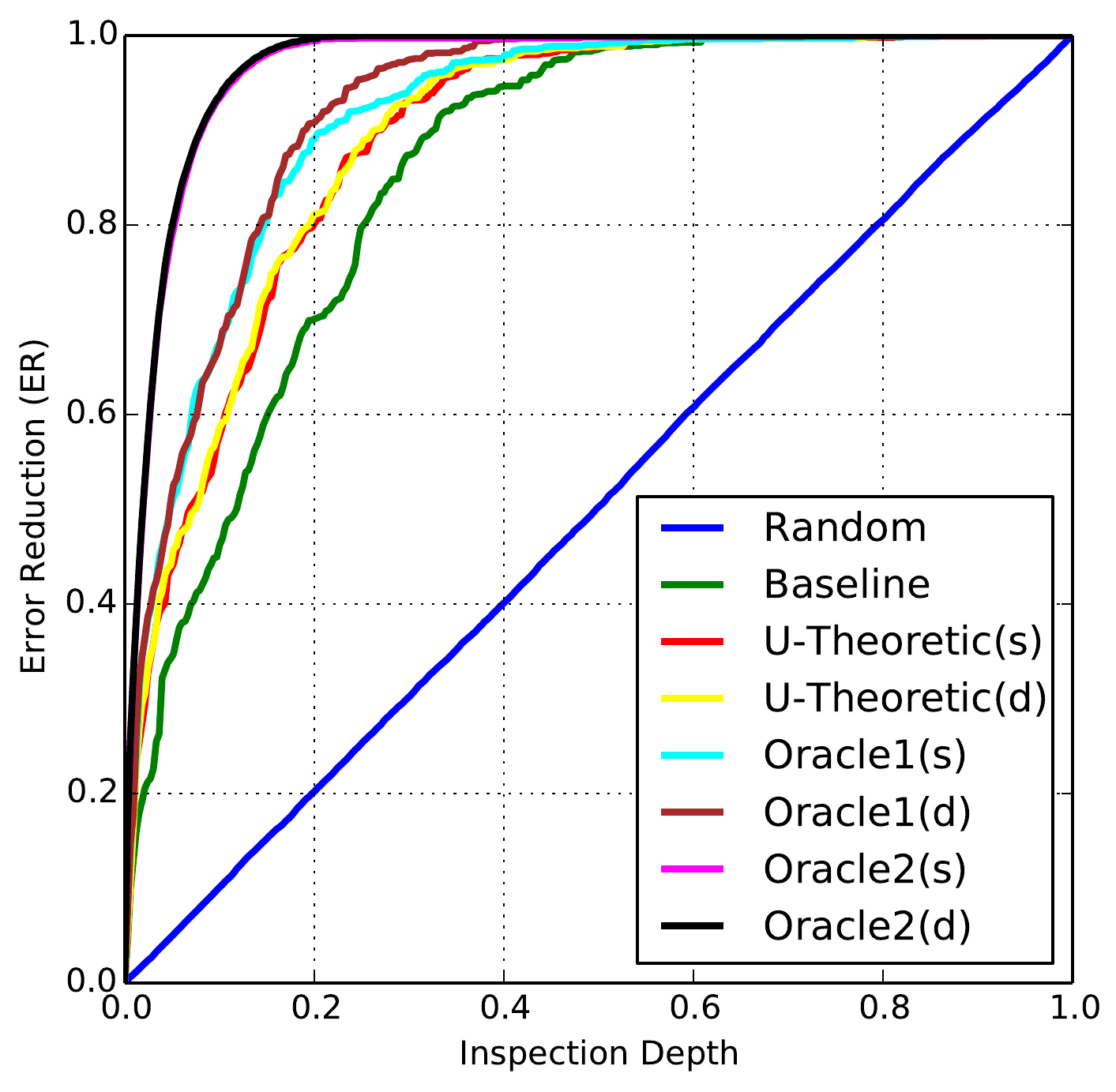}}
  \end{center} \caption{\label{fig:Reuters10}Results obtained by (a)
  splitting the \textsc{Reuters-21578} test set into 10 random,
  equally-sized parts, (b) running the analogous experiments of Figure
  \protect\ref{fig:Reuters} independently on each part, and (c)
  averaging the results across the 10 parts. The learners used are
  \mpb\ (left) and SVMs (right).}
\end{figure}


The results displayed in Figure \ref{fig:Reuters10} allow us to
visually appreciate that \textsf{U-Theoretic(s)} substantially
outperforms \textsf{Baseline} also in this context. This can be seen
also from Table \ref{tab:allthemacroresults}: for $\xi=.10$ the
relative improvement over \textsf{Baseline} is +110\% for \mpb\ and
+30\% for SVMs, and similarly substantial improvements are obtained
for the two other values of $\xi$ tested.


Incidentally, note that the \textsc{Reuters-21578/10} experiments
model an application scenario in which a set of automatically labelled
documents is split (e.g., to achieve faster throughput) among $10$
human annotators, each one entrusted with validating a part of the
set. In this case, each annotator is presented with a ranking of her
own document subset, and works exclusively on it\footnote{Actually, if
we did have $k$ annotators available, the best strategy would be to
generate the $k$ rankings in a ``round robin'' fashion, i.e., by
allotting to annotator $i$ the documents ranked (in the global
ranking) at the positions $r$ such that $(r \ \mathbf{mod} \ k) =
i$. This splitting method would guarantee that only the most promising
documents are validated by the annotators.}.


\subsubsection{Tiny test sets}\label{sec:tinytestsets}

\noindent In further experiments that we have run, we have split the
\textsc{Reuters-21578} test set even further, i.e., into 100
equally-sized parts of about 33 documents each, so as to test the
performance of Laplace smoothing methods in even more challenging
conditions. We call this experimental scenario
\textsc{Reuters-21578/100}. From an application point of view this is
a less interesting scenario than the two previously discussed ones,
since applying a ranking method to a set of 33 documents only is of
debatable utility, given that a human annotator confronted with the
task of validating just 33 documents can arguably check them all
without any need for ranking. The goal of these experiments is thus
checking whether our method can perform well even in extreme, albeit
scarcely realistic, conditions.

The detailed $ER_{\rho}^{M}(n)$ plots for this
\textsc{Reuters-21578/100} scenario are presented in Figure
\ref{fig:Reuters100}, while the $ENER_{\rho}^{M}$ results are reported
in Table \ref{tab:allthemacroresults}\footnote{From the next
experiments onwards, for reasons of space we will not include the full
plots in the style of Figures \ref{fig:Reuters} to
\ref{fig:Reuters100}, and will only report $ENER_{\rho}^{M}$
results.}.
\textsf{U-Theoretic(s)} still outperforms \textsf{Baseline}, with a
relative improvement of +42\% with \mpb\ and +21\% with SVMs with
$\xi=.10$, corresponding to $p=.696$; qualitatively similar
improvements are obtained with the other tested values of $\xi$.

\begin{figure}[t]
  \begin{center}
    \resizebox{.49\textwidth}{!}{\includegraphics{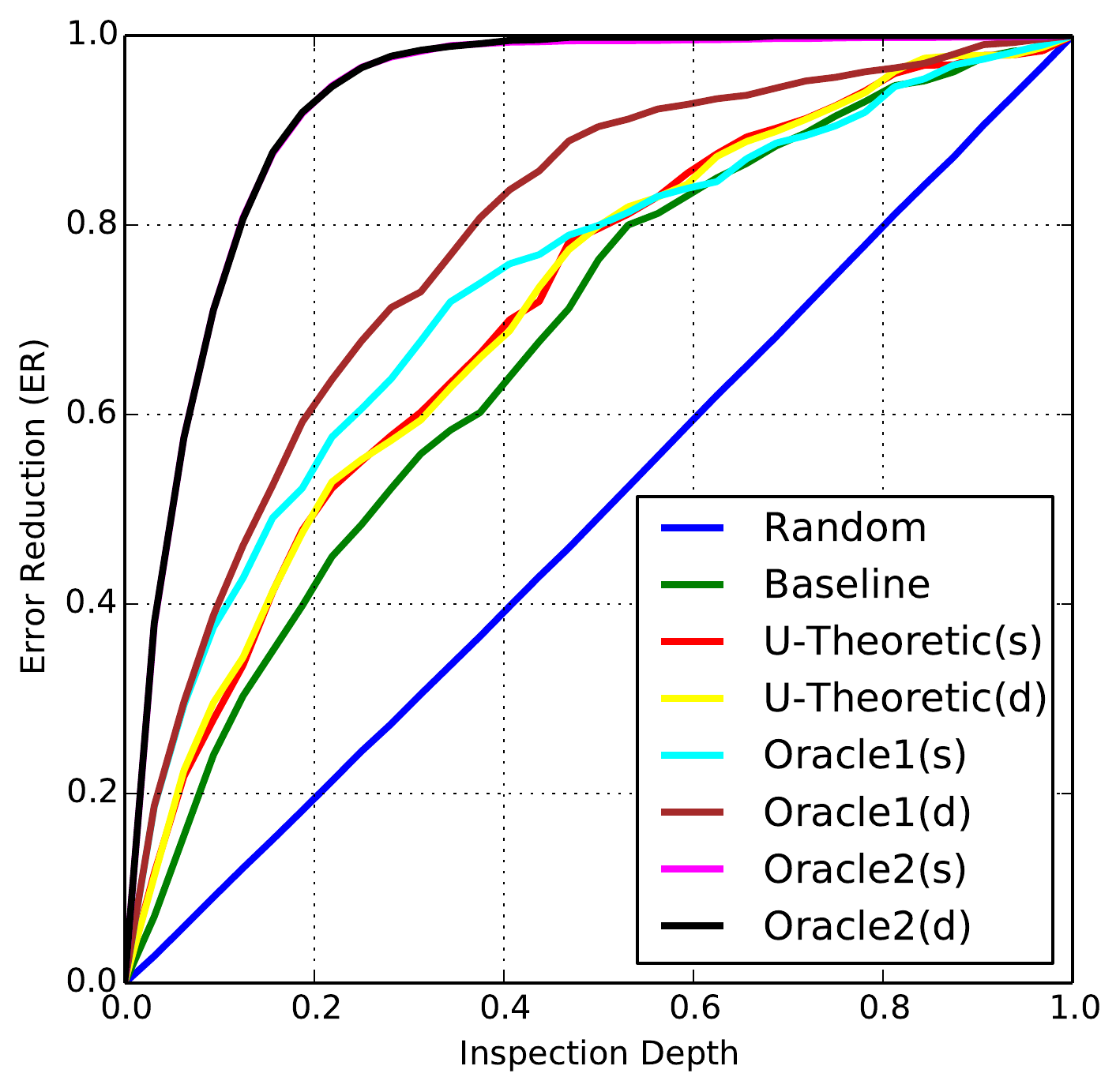}}
    \resizebox{.49\textwidth}{!}{\includegraphics{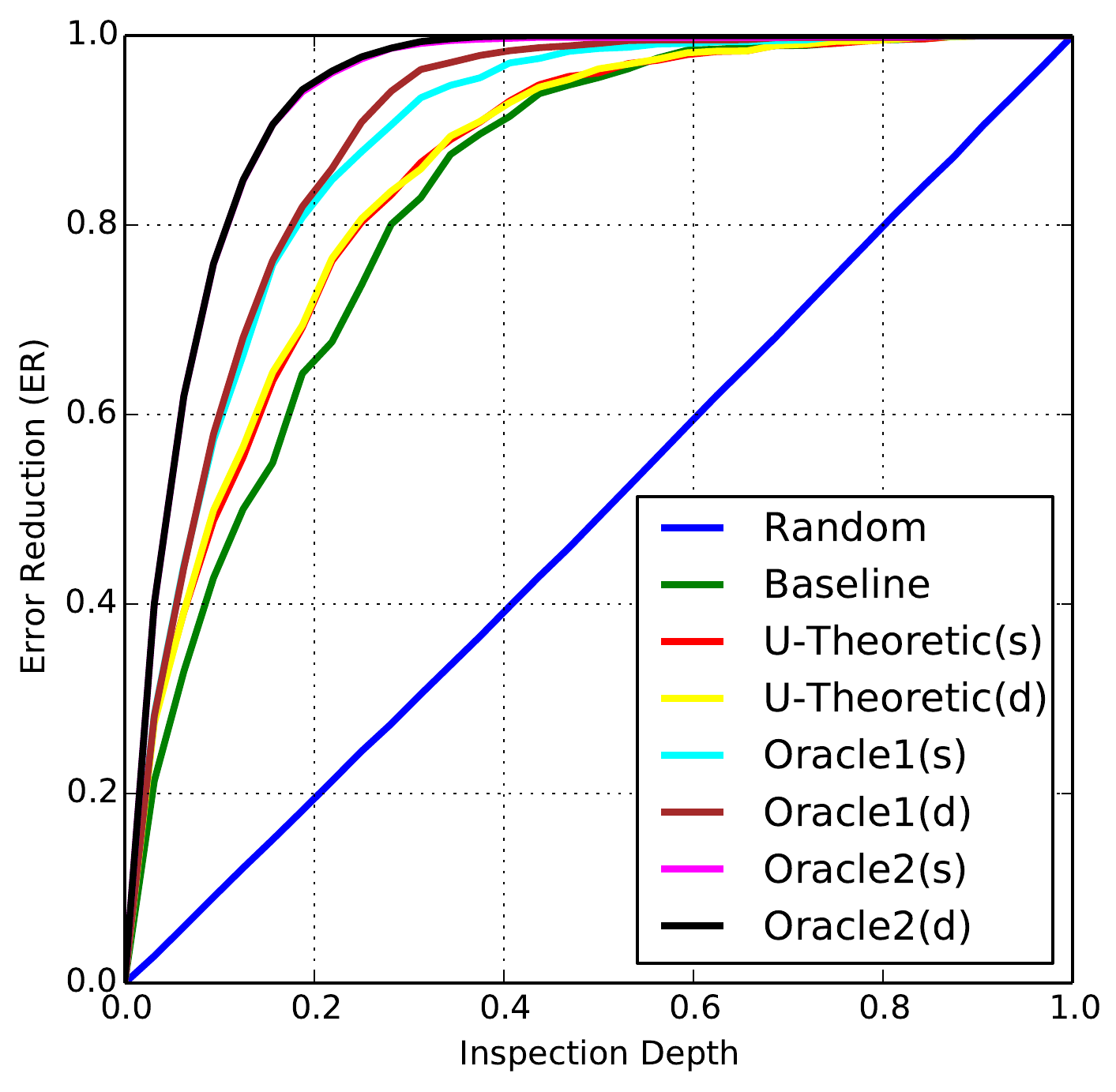}}
  \end{center} \caption{\label{fig:Reuters100}Same as Figure
  \ref{fig:Reuters10} but with \textsc{Reuters-21578/100} in place of
  \textsc{Reuters-21578/10}. The learners used are \mpb\ (left) and
  SVMs (right).}
\end{figure}
 
Note that in these experiments, unlike in those performed on the full
\textsc{Reuters-21578}, the \textsf{Oracle1(s)} method proves to be
markedly superior to \textsf{U-Theoretic(s)} (e.g., .247 vs.\ .172 in
terms of $ENER_{\rho}^{M}(.10)$ with \mpb, and similarly for other
values of $\xi$ and for the SVM learner); \blue{unlike in the previous two datasets, the difference between the two methods turns out to be statistically significant}. The reason is that, for a
smaller test set, (a) distribution drift is higher, (b) ``smoothing on
demand'' is invoked more frequently (because the likelihood that
contingency table cells have a value $\leq 1$ is higher), and (c) when
smoothing is indeed applied the distribution across the cells of the
contingency table is perturbed more strongly.

Note also that the $ER_{\rho}^{M}(n)$ curves are smoother than the
analogous curves for the full \textsc{Reuters-21578} and, although to
a lesser extent, those for \textsc{Reuters-21578/10}. This is due to
the fact that the curves in Figure \ref{fig:Reuters100} result from
averages across 100 different experiments, and the increase brought
about at rank $n$ is actually the average of the increases brought
about at rank $n$ in the 100 experiments.


\subsubsection{Large test sets}\label{sec:largetestsets}

\noindent While in the previous sections we have discussed experiments
on mid-sized to small (or very small) datasets, we now look at larger
datasets such as OHSUMED. The OHSUMED results in Table
\ref{tab:allthemacroresults} confirm the quality of
\textsf{U-Theoretic(s)}, which outperforms the purely confidence-based
baseline by +10\% (\mpb) and +9\% (SVMs) in terms of
$ENER_{\rho}^{M}(.10)$; qualitatively similar improvements are
obtained for the other two values of $\xi$ studied.

The \textsc{OHSUMED} collection is characterized by the presence of an
unusually large number (93.1\% of the entire lot) of unlabelled
documents (i.e., documents, that are negative examples for all
$c_{j}\in \mathcal{C}$) that originally belonged to other subtrees of
the MeSH tree. Since such a large percentage is unnatural, we have
generated (and also used in our experiments) a variant of
\textsc{OHSUMED} (called \textsc{OHSUMED-S}) by removing all the
unlabelled documents from both the training set and the test set.

As illustrated in Table \ref{tab:allthemacroresults}, on OHSUMED-S
\textsf{U-Theoretic(s)} outperforms the con\-fidence-based baseline by
a very large margin (+374\% with \mpb\ and +127\% with SVMs for
$\xi=.10$, with qualitatively similar results for the other two tested
values of $\xi$).


\subsubsection{Discussion}\label{sec:discussion}

\noindent In sum, the results discussed from Section
\ref{sec:midsizedtestsets} to the present one have unequivocally shown
that \textsf{U-Theoretic(s)} outperforms the confidence-based
baseline, usually by a large or very large margin, for all the five
tested datasets and for both tested learners.

Note that, for all five datasets and for both learners, the
improvements of the utility-theoretic methods over \textsf{Baseline}
are larger for smaller values of $\xi$. This indicates that the
difference between the two methods is larger for smaller validation
depths, i.e., where using the utility-theoretic method pays off the
most is at the very top of the ranking. This is an important feature
of this method, since it means that all human annotators, be they
persistent or not (i.e., independently of the depth at which they
validate), are going to benefit from this approach.


\section{An improved, ``dynamic'' ranking function for
SATC}\label{sec:rankingfunctiondynamic}

\noindent The utility-theoretic method discussed in Section
\ref{sec:rankingfunction} is reasonable but, in principle, suboptimal,
and its suboptimality derives from its ``static'' nature. To see this,
assume that the system has ranked the test documents according to the
strategy above, that the human annotator has started from the top of
the list and validated the labels of document $d_{i}$, that she has
found out that its label assignment for class $c_{j}$ is a false
negative, and that she has corrected it, thus bringing about an
increase in $F_{1}$ equivalent to
\begin{equation}
  \begin{aligned}
    \label{eq:pointwisegainpos}
    \frac{2(TP_{j}+1)}{2(TP_{j}+1)+FP_{j}+(FN_{j}-1)} - \
    \frac{2TP_{j}}{2TP_{j}+FP_{j}+FN_{j}}
  \end{aligned}
\end{equation}
\noindent Following this correction, the value of $FN_{j}$ is
decreased by 1 and the value of $TP_{j}$ is increased by 1. This means
that, when another false negative for $c_{j}$ is found and corrected,
the value of (\ref{eq:pointwisegainpos}) has changed.
In other words,
%
%
the improvement in $F_{1}$ due to the validation of a false negative
is not constant through the validation process. Of course, similar
considerations apply for false positives.

This suggests redefining the validation gains defined in Equations
\ref{eq:gainFP} and \ref{eq:gainFN} as
%
\begin{equation}
  \begin{aligned}
    \label{eq:pointwisegains}
    G(d_{i},fp_{j}) & = \frac{2TP_{j}}{2TP_{j}+(FP_{j}-1)+FN_{j}} - \frac{2TP_{j}}{2TP_{j}+FP_{j}+FN_{j}} \\
    G(d_{i},fn_{j}) & =
    \frac{2(TP_{j}+1)}{2(TP_{j}+1)+FP_{j}+(FN_{j}-1)} - \
    \frac{2TP_{j}}{2TP_{j}+FP_{j}+FN_{j}}
  \end{aligned}
\end{equation}
\noindent To see the novelty introduced with respect to Equation
\ref{eq:pointwisegains}, in the following we will discuss the case of
false negatives; the case of false positives is completely
analogous. The difference between Equation \ref{eq:gainFN} and
Equation \ref{eq:pointwisegains} is that the former equates
$G(d_{i},fn_{j})$ with the increase in $F_{1}(\hat\Phi_{j}(Te))$ that
would derive by correcting \emph{all of the documents in} $FN_{j}$
\emph{divided by their number}, while the latter equates
$G(d_{i},fn_{j})$ with the increase in $F_{1}(\hat\Phi_{j}(Te))$ that
would derive by correcting \emph{the next document in} $FN_{j}$. In
other words, we might say that Equation \ref{eq:gainFN} enforces the
notion of \emph{average gain}, while Equation \ref{eq:pointwisegains}
enforces the notion of \emph{pointwise gain}\footnote{\blue{Equations
\ref{eq:pointwisegains} might have also been formulated in a
continuous way, i.e., as partial derivatives of $F_{1}$ in the two
variables $TP_{j}$ and $TN_{j}$ (in other words, Equations
\ref{eq:pointwisegains} would thus represent the gradient of
$F_{1}$). We have preferred to stick to a discrete formulation, since
(a) Equations \ref{eq:gainFP} and \ref{eq:gainFN} are instead
\emph{not} naturally formulated as derivatives (exactly because they
represent average --} \blue{rather than pointwise -- gains), and since
(b) having Equations \ref{eq:gainFP}, \ref{eq:gainFN} and
\ref{eq:pointwisegains} all formulated in a common notation allows an
easier comparison among them.}}. The two versions return different
values of $G(d_{i},fn_{j})$: as the following example shows, it is
immediate to verify that if $FN_{j}$ contains more than one document,
the validation gains $G(d_{i},fn_{j})$ that derive by correcting
different documents are the same (by definition) if we use Equation
\ref{eq:gainFN} but are not the same if we use Equation
\ref{eq:pointwisegains}.

\begin{example}
  Suppose we have classified a set of 100 documents according to class
  $c_{j}$, and that the classification is such that $TP_{j}=10$,
  $FN_{j}=20$, $FP_{j}=30$, and $TN_{j}=40$. According to Equation
  \ref{eq:gainFN}, $G(d_{i},fn_{j})$ evaluates to $\approx 0.0190$ for
  each false negative corrected. Instead, according to Equation
  \ref{eq:pointwisegains}, $G(d_{i},fn_{j})$ evaluates to $\approx
  0.0241$ for the 1st false negative corrected, $\approx 0.0235$ for
  the 2nd, $\approx 0.0228$ for the 3rd, ..., down to $\approx 0.0147$
  for the 20th. \qed
\end{example}



\noindent Given this new definition we may implement a dynamic
strategy in which, instead of plainly sorting the test documents in
descending order of their $U(d_{i},\Omega)$ score, after each
correction is made we update $\hat{TP}_{j}^{La}$, $\hat{FP}_{j}^{La}$,
$\hat{FN}_{j}^{La}$ by adding and subtracting 1 where appropriate, we
recompute $G(d_{i},fp_{j})$, $G(d_{i},fn_{j})$ and $U(d_{i},\Omega)$,
and we use the newly computed $U(d_{i},\Omega)$ values when selecting
the document that should be presented next to the human annotator.
In detail, the following steps are iteratively performed:
\begin{enumerate}

\item For all classes $c_{j}\in \mathcal{C}$, compute
  $G(d_{i},fp_{j})$ and/or $G(d_{i},fn_{j})$ using Equations
  \ref{eq:pointwisegains};

\item \label{item:identifymax} If the human annotator does not want to
  stop validating documents, then identify the document $d_{max}\equiv
  \arg\displaystyle\max_{d_{i}\in Te} U(d_{i},\Omega)$ for which total
  utility is maximised;
  
  
\item Remove $d_{max}$ from $Te$;
 
\item \label{item:validatelabels} For all $c_{j}\in \mathcal{C}$, have
  the human annotator check the label attached by $\hat\Phi_{j}$ to
  $d_{max}$; if all these labels are correct
  go to Step \ref{item:identifymax}; else, for all classes $c_{j}\in
  \mathcal{C}$ for which the label attached by $\hat\Phi_{j}$ to
  $d_{max}$ is incorrect:
 
  \begin{enumerate}

  \item Have the human annotator correct the label;
 
  \item If $d_{max}$ was a false positive for $c_{j}$, decrease
    $\hat{FP}_{j}^{La}$ by 1; if it was a false negative for $c_{j}$,
    increase $\hat{TP}_{j}^{La}$ by 1 and decrease $\hat{FN}_{j}^{La}$
    by 1;
      
  \item \label{item:resmooth} Re-smooth $\hat{TP}_{j}^{La}$,
    $\hat{FP}_{j}^{La}$, $\hat{FN}_{j}^{La}$ if needed;

  \item \label{item:recomputeG} Recompute $G(d_{i},fp_{j})$ and/or
    $G(d_{i},fn_{j})$ and go back to Step \ref{item:identifymax}.
 

  \end{enumerate}
 
 
\end{enumerate}
\noindent This might also be dubbed an \emph{incremental} ranking
strategy, in the sense pioneered in \cite{Aalbersberg:1992fk} for
relevance feedback in ad-hoc search, in the sense that the values of
$G(d_{i},fp_{j})$ and $G(d_{i},fn_{j})$ are incrementally updated so
that the $U(d_{i},\Omega)$ function reflects the fact that part of
$Te$ has indeed been corrected. In keeping with \cite{Brandt:2011fk}
we prefer to call it a \emph{dynamic} strategy, and to call the one of
Section \ref{sec:rankingfunction} a \emph{static} one.

Note that in Step \ref{item:identifymax} we simply compute the maximum
element (according to $U(d_{i},\Omega)$) of $Te$ instead of sorting
the entire set, since we can perform this step in $O(|Te|)$ instead of
$O(|Te|\log|Te|)$\footnote{When computing this maximum element returns
repeatedly a document whose labels are all correct, the lack of a
sorting step entails the need of computing the maximum element several
times in a row with the values of $G(d_{i},fp_{j})$ and
$G(d_{i},fn_{j})$ unchanged. In these cases, the presence of a sorting
step would thus have been advantageous. However, the likelihood that
this situation occurs tends to be small, especially when
$|\mathcal{C}|$ is large, thus making the computation of the maximum
element preferable to sorting.}. Furthermore, note that in this
algorithm the re-computation of $U_{j}(d_{i},\Omega)$ does \emph{not}
entail the recomputation of the probabilities $P(fp_{j})$ and/or
$P(fn_{j})$ of Equation \ref{eq:ourexpectedutility}, since these
probabilities are computed (i.e., calibrated) once for all,
immediately after the training phase.

Note also that computing validation gains via Equations
\ref{eq:gainFP} and \ref{eq:gainFN} is the only possibility within the
static method (since the values of $G(d_{i},fp_{j})$ and
$G(d_{i},fn_{j})$ produced must be used unchanged throughout the
process), but is clearly inadequate in a dynamic context, in which
validation gains are always supposed to be up-to-date reflections of
the current situation.


The dynamic nature of this method makes it clear why, as specified at
the end of Section \ref{sec:smoothing}, we smooth the cell count
estimates only ``on demand'' (see also Step \ref{item:resmooth} of the
above algorithm), i.e., only if any of $\hat{TP}_{j}^{ML}$,
$\hat{FP}_{j}^{ML}$, $\hat{FN}_{j}^{ML}$ is $< 1$. To see this,
suppose that we smooth $\hat{TP}_{j}^{ML}$, $\hat{FP}_{j}^{ML}$,
$\hat{FN}_{j}^{ML}$ at each iteration, even when not strictly
needed. Adding a count of one to each of them at each iteration means
that, after $k$ iterations, $k$ counts have been added to each of
them; this means that, after many iterations, the counts added to the
cells have completely disrupted the relative proportions among the
cells that result from the maximum-likelihood estimation. This would
likely make the dynamic method underperform the static method, which
does not suffer from this problem since the maximum-likelihood
estimates are smoothed only once. As a result, we smooth a contingency
table only when strictly needed, i.e., when one of
$\hat{TP}_{j}^{ML}$, $\hat{FP}_{j}^{ML}$, $\hat{FN}_{j}^{ML}$ is $<1$.



By solving the inequality $G(d_{i},fn_{j})>G(d_{i},fp_{j})$ we may
find out under which conditions correcting a false negative yields a
higher gain than correcting a false positive. It turns out that, when
validation gains are defined according to Equation
\ref{eq:pointwisegains}, $G(d_{i},fn_{j})>G(d_{i},fp_{j})$ whenever
$FN+FP>1$, i.e., practically always. Of course, this need not be the
case for evaluation functions different from $F_{1}$, and in
particular for instances of $F_{\beta}$ with $\beta\not = 1$.

%
From the standpoint of total computational cost, our dynamic technique
is $O(|Te|\cdot (|\mathcal{C}|+|Te|))$, since (i) computing the
$U(d_{i},\Omega)$ score for $|Te|$ documents and computing their
maximum according to the computed $U(d_{i},\Omega)$ score can be done
in $O(|Te|\cdot|\mathcal{C}|)$ steps, and (ii) this step must be
repeated $O(|Te|)$ times.
This policy is thus, as expected, computationally more expensive than
the previous one.


\subsection{Experiments}
\label{sec:experimentsdynamic}

\noindent The results of the experiments with the dynamic version of
our utility-theoretic method and of our two oracle-based methods are
reported in Figures \ref{fig:Reuters} to \ref{fig:Reuters100} and in
Table \ref{tab:allthemacroresults}, where they are indicated as
\textsf{U-Theoretic(d)}, \textsf{Oracle1(d)} and
\textsf{Oracle2(d)}. Of course there exists no dynamic version of the
baseline method, since this latter does not involve validation gains.

The first observation that can be drawn from these results is the fact
that \textsf{U-Theoretic(d)} is not superior to
\textsf{U-Theoretic(s)}, as could instead have been expected. In fact,
in Figures \ref{fig:Reuters} to \ref{fig:Reuters100} the curves
corresponding to the former are barely distinguishable from those
corresponding to the latter, and the numeric results reported in Table
\ref{tab:allthemacroresults} show no substantial difference
either; \blue{as reported in Table \ref{tab:stat}, in 7 out of 10 cases (2 learners $\times$ 5 datasets) the difference is not statistically significant}. Note that there are extremely small differences also between
\textsf{Oracle1(s)} and \textsf{Oracle1(d)}; \blue{again, in 7 out of 10 cases no statistically significant difference can be detected.} This shows that the lack
of any substantial difference between static and dynamic is not due to
a possible suboptimality of the method for estimating contingency
table cells (including the method adopted for smoothing the
estimates). Analogously, note also the extremely small differences
between \textsf{Oracle2(s)} and \textsf{Oracle2(d)} \blue{(again, no statistically significant difference in 7 out of 10 cases)}, which indicates
that the culprit is not the method for estimating the probabilities of
misclassification.

This substantial equivalence between the static and the dynamic
methods is somehow surprising, since on a purely intuitive basis the
dynamic method seems definitely superior to the static one. We think
that the reason for this apparently counterintuitive results is that,
when validation gains are recomputed in Step \ref{item:recomputeG} of
the algorithm, the magnitude of the update (i.e., the difference
between validation gains before and after the update) is too small to
make an impact. This is especially true for large test sets, where
incrementing or decrementing by 1 the value of a contingency cell
makes too tiny a difference, since that value is very large.

Actually, the part of Figure \ref{fig:Reuters} relative to \mpb\
displays an apparently strange phenomenon, i.e., the fact that for
some values of $\xi$ the \textsf{Oracle2(s)} method outperforms
\textsf{Oracle2(d)}. A similar phenomenon can be noticed in some of
the cells of Table \ref{tab:allthemacroresults}, where the static
version of either \textsf{Oracle1} or \textsf{Oracle2} outperforms,
even if by a small margin, the dynamic version. This seems especially
strange for \textsf{Oracle2(d)}, which is the theoretically optimal
method (since it is a method that operates with perfect
foreknowledge), and as such should be impossible to beat. The reason
for this apparently counterintuitive behaviour lies not in the ranking
methods, but in a counterintuitive property of $F_{1}$, i.e., the fact
that, when $TP=FN=0$ (i.e., there are no positives in the gold
standard -- and 25 out of 115 classes in the dataset used in Figure
\ref{fig:Reuters} have this property), its value is 0 when $FP>0$ but
1 when $FP=0$ (so, $TP=FP=FN=0$ is a ``point of discontinuity'' for
$F_{1}$). This essentially means that, when $TP=FN=0$ and $FP>0$,
$G(d_{i},fn_{j})$ is $1/|FP|$ for the static method and 0 for the
dynamic method; i.e., in this case the dynamic method does not provide
any incentive for correcting a false positive, while the static method
does. As a result, the static method can speed up the correction of
false positives more than the dynamic method does. As mentioned above,
this phenomenon exposes a suboptimality not of the dynamic method, but
of the $F_{1}$ function.

In Table \ref{tab:computationtimes} we report the actual computation
times incurred by both \textsf{U-Theoretic(s)} and
\textsf{U-Theoretic(d)} on our five datasets\footnote{The times
reported are relative to an experiment in which the entire test set is
validated; this is because, in a simulated experiment, the entire test
set must be validated in order to compute the $ER_{\rho}^{M}(n)$
values reported in Figures \ref{fig:Reuters} to
\ref{fig:Reuters100}. In a realistic setting in which only a portion
of the ranked list is validated, the difference between
\textsf{U-Theoretic(s)} and \textsf{U-Theoretic(d)} is smaller, since
the cost of recomputing validation gains is roughly proportional to
the validation depth, and since this cost affects
\textsf{U-Theoretic(d)} but not \textsf{U-Theoretic(s)}.}. These
figures confirm that the dynamic method is (as already discussed
above) substantially more expensive to run than the static method; in
particular, the magnitude of this difference, together with the
marginal (if any) accuracy improvements brought about by the dynamic
method over the static one, shows that the static method is much more
cost-effective than the dynamic one. In other words, the bad news is
that the dynamic method brings about no improvement; the good news is
that the computationally cheaper static method is hard to beat.

\begin{table}[t]
  \begin{center}
    \tbl{
    \label{tab:computationtimes}Comparison between the actual computation times (in
    seconds) of the \textsf{U-Theoretic(s)} and
    \textsf{U-Theoretic(d)} methods on our five datasets.}{
    \begin{tabular}{|c|c||c|c|}
      \hline
      Dataset & Method & \mpb\ & \hspace{1em} SVMs \hspace{1em} \\
      \hline\hline
      \multirow{2}{*}{\textsc{Reuters-21578}}     & \textsf{U-Theoretic(s)} & \hspace{1em}0.426 & \hspace{1em}0.452 \\
      & \textsf{U-Theoretic(d)} & \hspace{1em}3.128 & \hspace{1em}3.021\\
      \hline
      \multirow{2}{*}{\textsc{Reuters-21578/10}}  & \textsf{U-Theoretic(s)} & \hspace{1em}0.166 & \hspace{1em}0.153 \\
      & \textsf{U-Theoretic(d)} & \hspace{1em}0.195 & \hspace{1em}0.198 \\
      \hline
      \multirow{2}{*}{\textsc{Reuters-21578/100}} & \textsf{U-Theoretic(s)} & \hspace{1em}0.033 & \hspace{1em}0.033 \\
      & \textsf{U-Theoretic(d)} & \hspace{1em}0.046 & \hspace{1em}0.044 \\
      \hline
      \multirow{2}{*}{\textsc{OHSUMED}}           & \textsf{U-Theoretic(s)} & \hspace{.5em}10.282 & \hspace{.5em}11.251 \\
      & \textsf{U-Theoretic(d)} & 500.047 & 577.864 \\
      \hline
      \multirow{2}{*}{\textsc{OHSUMED-S}}         & \textsf{U-Theoretic(s)} & \hspace{1em}0.418 & \hspace{1em}0.424 \\
      & \textsf{U-Theoretic(d)} & \hspace{1em}4.731 & \hspace{1em}4.195 \\                                        
      \hline
    \end{tabular}
    }
  \end{center}
\end{table}

\section{A ``micro-oriented'' ranking function for SATC}
\label{sec:microranking}

\noindent In Section \ref{sec:definitions} we have assumed that the
evaluation of classification algorithms across the $|\mathcal{C}|$
classes of interest is performed by \emph{macro-averaging} the $F_{1}$
results obtained for the individual classes
$c_{j}\in\mathcal{C}$. Consistently with this view, in Section
\ref{sec:effmeasures} we have introduced macro-averaged versions of
$E_{1}$, $ER_{\rho}$, $NER_{\rho}$, and $ENER_{\rho}$. macro-averaging
across the classes in $|\mathcal{C}|$ essentially means paying equal
attention to all of them, irrespective of their frequency or other
such characteristics.

However, there is an alternative, equally important way to evaluate
effectiveness when a set of $|\mathcal{C}|$ classes is involved,
namely, \emph{micro-averaged} effectiveness. While macro-averaged
measures are computed by first computing the measure of interest
individually on each class-specific contingency table and then
averaging the results, micro-averaged measures are computed by merging
the $|\mathcal{C}|$ contingency tables into a single one (via summing
the values of the corresponding cells) and then computing the measure
of interest on the resulting table. For instance, micro-averaged
$F_{1}$ (noted $F_{1}^{\mu}$) is obtained by (i) computing the
category-specific values $TP_{j}$, $FP_{j}$ and $FN_{j}$ for all
$c_{j}\in\mathcal{C}$, (ii) obtaining $TP$ as the sum of the
$TP_{j}$'s (same for $FP$ and $FN$), and then (iii) applying Equation
\ref{eq:F1}. Measures such as $E_{1}^{\mu}$, $ER_{\rho}^{\mu}$,
$NER_{\rho}^{\mu}$, and $ENER_{\rho}^{\mu}$ are defined in the obvious
way.  The net effect of using a single, global contingency table is
that micro-averaged measures pay more attention to more frequent
classes, i.e., the more the members of a class $c_{j}$ in the test
set, the more the measure is influenced by $c_{j}$.

Neither macro- nor micro-averaging are the ``right'' way to average in
evaluating multi-label multi-class classification; it is instead the
case that in some applications we may want to pay equal attention to
all the classes (in which case macro-averaging would be our evaluation
method of choice), while in some other applications we may want to pay
more attention to the most frequent classes (in which case we should
opt for micro-averaging).

While we have not explicitly discussed this, the method of Section
\ref{sec:rankingfunction} was devised with macro-averaged
effectiveness in mind. To see this, note that the $U(d_{i},\Omega)$
function of Equation \ref{eq:utilitysum} is based on an unweighted sum
of the class-specific $U_{j}(d_{i},\Omega)$ scores, i.e., it pays
equal importance to all classes in $\mathcal{C}$. This means that
Equation \ref{eq:utilitysum} is optimized for metrics that also pay
equal attention to all classes, as all macro-averaged measures do.  We
now describe a way to modify the method of Section
\ref{sec:rankingfunction} in such a way that it is instead optimal
when our effectiveness measure of choice (e.g., $ENER_{\rho}$) is
micro-averaged.
To do this,
%
we do away with Equation \ref{eq:utilitysum} and (similarly to what
happens for $F_{1}^{\mu}$ and $E_{1}^{\mu}$) compute instead
$U(d_{i},\Omega)$ directly on a single, global contingency table
obtained by the cell-wise sum of the class-specific contingency
tables. That is, we redefine $U(d_{i},\Omega)$ as
%
%
\begin{equation}
  \label{eq:utilitymicro}
  U(d_{i}, \Omega)=\sum_{c_{j}\in \mathcal{C}}\sum_{\omega_{k}\in\{tp_{j},fp_{j},fn_{j},tn_{j}\}}P(\omega_{k})G(d_{i},\omega_{k})
\end{equation}
%
where
%
\begin{equation}
  \begin{aligned}
    \label{eq:gainmicro}
    G(d_{i},fp_{j}) & = \frac{1}{FP} (F_{1}^{FP}(\hat\Phi(Te))-F_{1}(\hat\Phi(Te))) \\
    & = \frac{1}{FP} (\frac{2TP}{2TP+FN} - \
    \frac{2TP}{2TP+FP+FN}) \\
    G(d_{i},fn_{j}) & = \frac{1}{FN}
    (F_{1}^{FN}(\hat\Phi(Te))-F_{1}(\hat\Phi(Te))) \\ & = \frac{1}{FN}
    (\frac{2(TP+FN)}{2(TP+FN)+FP} - \ \frac{2TP}{2TP+FP+FN})
  \end{aligned}
\end{equation}
\noindent Equations \ref{eq:gainmicro} are the same as Equation
\ref{eq:gainFP} and \ref{eq:gainFN}, but for the fact that the latter
are class-specific (as indicated by the index $j$) while the former
are global. This is due to the fact that, when using micro-averaging,
there is a single contingency table, and the gain obtained by
correcting, say, a false positive for $c_{x}$ is equal to the gain
obtained by correcting a false positive for $c_{y}$, for any $c_{x},
c_{y}\in \mathcal{C}$. Of course, Equations \ref{eq:gainmicro} are to
be applied when the \emph{static} method of Section
\ref{sec:rankingfunction} needs to be optimized for micro-averaging;
when we instead want to do the same optimization for the
\emph{dynamic} method of Section \ref{sec:rankingfunctiondynamic}, we
need instead to apply, in the obvious way, ``global'' versions of
Equations \ref{eq:pointwisegains}.

Actually, a second aspect in the method of Section
\ref{sec:rankingfunction} that we need to change in order for it to be
optimized for micro-averaging is the probability calibration method
discussed in Section \ref{sec:calib}. In fact, Equation
\ref{sec:plattcalibration} is clearly devised with macro-averaging in
mind, since it minimizes \emph{the average across the} $c_{j}\in
\mathcal{C}$ of the difference between the number $Pos_{j}^{Tr}$ and
the expected number $\mathrm{E}[Pos_{j}^{Tr}]$ of positive training
examples of class $c_{j}$. Again, all classes are given equal
attention. For our micro-averaging-oriented method we thus replace
Equation \ref{sec:plattcalibration} with
\begin{equation}
  \begin{aligned}
    \label{sec:plattcalibrationformicro}
    & \arg\min_{\sigma}|Pos^{Tr}-\mathrm{E}[Pos^{Tr}]|  = \\
    & \arg\min_{\sigma}|\displaystyle\sum_{c_{j}\in \mathcal{C}}Pos_{j}^{Tr}-\displaystyle\sum_{c_{j}\in \mathcal{C}}\mathrm{E}[Pos^{Tr}_{j}]|  = \\
    & \arg\min_{\sigma}|\displaystyle\sum_{c_{j}\in \mathcal{C}}Pos^{Tr}_{j}-\displaystyle\sum_{c_{j}\in \mathcal{C}}\sum_{d_{i}\in Tr}P(c_{j}|d_{i})|  = \\
    & \arg\min_{\sigma}|\displaystyle\sum_{c_{j}\in
    \mathcal{C}}Pos^{Tr}_{j}-\displaystyle\sum_{c_{j}\in
    \mathcal{C}}\sum_{d_{i}\in
    Tr}\frac{e^{\sigma\hat\Phi_{j}(d_{i})}}{e^{\sigma\hat\Phi_{j}(d_{i})}+1}|
  \end{aligned}
\end{equation}
\noindent where the difference between the number and the expected
number of training examples \emph{in the global contingency table} is
minimized. It is easy to verify that the two methods may return
different values of $\sigma$, as the following example shows.
\begin{example}
  Suppose that $\mathcal{C}=\{c_{1},c_{2}\}$, that $Pos_{1}^{Tr}=20$
  and that $Pos_{2}^{Tr}=10$. Suppose that when $\sigma=a$ then
  $\mathrm{E}[Pos_{1}^{Tr}]=18$ and $\mathrm{E}[Pos_{2}^{Tr}]=8$,
  while when $\sigma=b$ then $\mathrm{E}[Pos_{1}^{Tr}]=17$ and
  $\mathrm{E}[Pos_{2}^{Tr}]=13$. According to Equation
  \ref{sec:plattcalibration} value $a$ is better than $b$ (since
  $\frac{1}{|\mathcal{C}|}\sum_{c_{j}\in
  \mathcal{C}}|Pos^{Tr}_{j}-\mathrm{E}[Pos^{Tr}_{j}]|$ is equal to 2
  for $\sigma=a$ and to 3 for $\sigma=b$), but according to Equation
  \ref{sec:plattcalibrationformicro} value $b$ is better than $a$
  (since $|Pos^{Tr}-\mathrm{E}[Pos^{Tr}]|$ is equal to 4 for
  $\sigma=a$ and to 0 for $\sigma=b$). \qed
\end{example}

\noindent The same smoothing methods as discussed in Section
\ref{sec:smoothing} can instead be used; however, note that smoothing
is likely to be needed much less frequently (if at all) here since,
given that we now have a single global contingency table, it is much
less likely that any of its cells have values $<1$.


\subsection{Experiments}
\label{sec:experimentsmicro}

\noindent The experiments with our ``micro-oriented'' methods are
reported in Table \ref{tab:allthemicroresults}. Note that, since the
method we use as baseline corresponds (as noted in Section
\ref{sec:bounds}) to using \textsf{U-Theoretic(s)} with all validation
gains set to 1, the baseline we use here is different from the
baseline we had used in Section \ref{sec:results}, since the latter
was optimized for macro-averaging while the one we use here is
optimized for micro-averaging. This guarantees that, in both cases,
our baselines are strong ones.

\begin{table}[h]
  \begin{center} 
  \tbl{\label{tab:allthemicroresults}As Table \ref{tab:allthemacroresults}, but with $ENER_{\rho}^{\mu}(\xi)$ in place of $ENER_{\rho}^{M}(\xi)$.}{
    \resizebox{\textwidth}{!} {
    \begin{tabular}{|c|l||rr|rr|rr||rr|rr|rr|}
      \hline & & \multicolumn{6}{c||}{\mpb} & \multicolumn{6}{c|}{SVMs} \\
      \hline 
      & &
      \multicolumn{2}{c|}{$\xi=0.05$} &
      \multicolumn{2}{c|}{$\xi=0.10$} &
      \multicolumn{2}{c||}{$\xi=0.20$} &
      \multicolumn{2}{c|}{$\xi=0.05$} &
      \multicolumn{2}{c|}{$\xi=0.10$} &
      \multicolumn{2}{c|}{$\xi=0.20$} \\
      \hline\hline
      \multirow{7}{*}{\begin{sideways}{\textsc{Reuters-21578}}\end{sideways}}
      & Baseline & .107 && .167 && .222 && .240 && .325 && .389 & \\
      \cline{2-14}
      & U-Theoretic(s) & .107 & (+0\%) & .168 & (+1\%) & .224 & (+1\%) & .246 & (+3\%) & .332 & (+2\%) & .395 & (+2\%)  \\
      \cline{2-14}
      & U-Theoretic(d) & .107 & (+0\%) & .167 & (+0\%) & .224 & (+1\%) & .246 & (+3\%) & .331 & (+2\%) & .394 & (+1\%)  \\
      \cline{2-14}
      & Oracle1(s) & .107 & (+0\%) & .168 & (+1\%) & .224 & (+1\%) & .246 & (+3\%) & .332 & (+2\%) & .395 & (+2\%)  \\
      \cline{2-14}
      & Oracle1(d) & .107 & (+0\%) & .167 & (+0\%) & .224 & (+1\%) & .246 & (+3\%) & .331 & (+2\%) & .395 & (+2\%)  \\
      \cline{2-14}
      & Oracle2(s) & .333 & (+211\%) & .448 & (+168\%) & .512 & (+131\%) & .394 & (+64\%) & .506 & (+56\%) & .556 & (+43\%)  \\
      \cline{2-14}
      & Oracle2(d) & .333 & (+211\%) & .448 & (+168\%) & .512 & (+131\%) & .394 & (+64\%) & .506 & (+56\%) & .556 & (+43\%)  \\
      \hline\hline
      \multirow{7}{*}{\begin{sideways}{\textsc{Reuters-21578/10}}\end{sideways}}
      & Baseline & .110 && .169 && .222 && .232 && .317 && .380 & \\
      \cline{2-14}
      & U-Theoretic(s) & .112 & (+2\%) & .171 & (+1\%) & .224 & (+1\%) & .237 & (+2\%) & .323 & (+2\%) & .386 & (+2\%)  \\
      \cline{2-14}
      & U-Theoretic(d) & .113 & (+3\%) & .171 & (+1\%) & .224 & (+1\%) & .238 & (+3\%) & .322 & (+2\%) & .383 & (+1\%)  \\
      \cline{2-14}
      & Oracle1(s) & .112 & (+2\%) & .171 & (+1\%) & .224 & (+1\%) & .237 & (+2\%) & .324 & (+2\%) & .386 & (+2\%)  \\
      \cline{2-14}
      & Oracle1(d) & .113 & (+3\%) & .171 & (+1\%) & .224 & (+1\%) & .238 & (+3\%) & .324 & (+2\%) & .386 & (+2\%)  \\
      \cline{2-14}
      & Oracle2(s) & .325 & (+195\%) & .438 & (+159\%) & .502 & (+126\%) & .385 & (+66\%) & .496 & (+56\%) & .547 & (+44\%)  \\
      \cline{2-14}
      & Oracle2(d) & .325 & (+195\%) & .438 & (+159\%) & .502 & (+126\%) & .385 & (+66\%) & .496 & (+56\%) & .547 & (+44\%)  \\
      \hline\hline
      \multirow{7}{*}{\begin{sideways}{\textsc{Reuters-21578/100}}\end{sideways}}
      & Baseline & .102 && .158 && .208 && .223 && .301 && .361 & \\
      \cline{2-14}
      & U-Theoretic(s) & .107 & (+5\%) & .163 & (+3\%) & .212 & (+2\%) & .224 & (+0\%) & .305 & (+1\%) & .366 & (+1\%)  \\
      \cline{2-14}
      & U-Theoretic(d) & .106 & (+4\%) & .162 & (+3\%) & .211 & (+1\%) & .226 & (+1\%) & .304 & (+1\%) & .363 & (+1\%)  \\
      \cline{2-14}
      & Oracle1(s) & .115 & (+13\%) & .170 & (+8\%) & .216 & (+4\%) & .232 & (+4\%) & .317 & (+5\%) & .377 & (+4\%)  \\
      \cline{2-14}
      & Oracle1(d) & .116 & (+14\%) & .170 & (+8\%) & .217 & (+4\%) & .235 & (+5\%) & .322 & (+7\%) & .383 & (+6\%)  \\
      \cline{2-14}
      & Oracle2(s) & .318 & (+212\%) & .429 & (+172\%) & .492 & (+137\%) & .367 & (+65\%) & .481 & (+60\%) & .534 & (+48\%)  \\
      \cline{2-14}
      & Oracle2(d) & .318 & (+212\%) & .429 & (+172\%) & .492 & (+137\%) & .367 & (+65\%) & .481 & (+60\%) & .534 & (+48\%)  \\
      \hline\hline
      \multirow{7}{*}{\begin{sideways}{\textsc{OHSUMED}}\end{sideways}}
      & Baseline & .442 && .552 && .583 && .492 && .600 && .620 & \\
      \cline{2-14}
      & U-Theoretic(s) & .440 & (+0\%) & .549 & (-1\%) & .580 & (-1\%) & .496 & (+1\%) & .602 & (+0\%) & .621 & (+0\%)  \\
      \cline{2-14}
      & U-Theoretic(d) & .442 & (+0\%) & .552 & (+0\%) & .582 & (+0\%) & .496 & (+1\%) & .602 & (+0\%) & .621 & (+0\%)  \\
      \cline{2-14}
      & Oracle1(s) & .439 & (-1\%) & .549 & (-1\%) & .580 & (-1\%) & .497 & (+1\%) & .602 & (+0\%) & .621 & (+0\%)  \\
      \cline{2-14}
      & Oracle1(d) & .441 & (+0\%) & .551 & (+0\%) & .582 & (+0\%) & .497 & (+1\%) & .603 & (+1\%) & .621 & (+0\%)  \\
      \cline{2-14}
      & Oracle2(s) & .660 & (+49\%) & .733 & (+33\%) & .711 & (+22\%) & .704 & (+43\%) & .761 & (+27\%) & .727 & (+17\%)  \\
      \cline{2-14}
      & Oracle2(d) & .660 & (+49\%) & .733 & (+33\%) & .711 & (+22\%) & .704 & (+43\%) & .761 & (+27\%) & .727 & (+17\%)  \\
      \hline\hline
      \multirow{7}{*}{\begin{sideways}{\textsc{OHSUMED-S}}\end{sideways}}
      & Baseline & .044 && .068 && .094 && .058 && .096 && .136 & \\
      \cline{2-14}
      & U-Theoretic(s) & .044 & (+1\%) & .069 & (+3\%) & .096 & (+2\%) & .063 & (+10\%) & .102 & (+7\%) & .143 & (+5\%)  \\
      \cline{2-14}
      & U-Theoretic(d) & .044 & (+1\%) & .070 & (+3\%) & .097 & (+3\%) & .066 & (+14\%) & .104 & (+9\%) & .144 & (+6\%)  \\
      \cline{2-14}
      & Oracle1(s) & .044 & (+1\%) & .069 & (+3\%) & .096 & (+2\%) & .064 & (+10\%) & .103 & (+8\%) & .143 & (+5\%)  \\
      \cline{2-14}
      & Oracle1(d) & .044 & (+1\%) & .070 & (+3\%) & .097 & (+3\%) & .066 & (+15\%) & .105 & (+10\%) & .144 & (+6\%)  \\
      \cline{2-14}
      & Oracle2(s) & .149 & (+242\%) & .221 & (+227\%) & .287 & (+205\%) & .175 & (+203\%) & .259 & (+171\%) & .330 & (+143\%)  \\
      \cline{2-14}
      & Oracle2(d) & .149 & (+242\%) & .221 & (+227\%) & .287 & (+205\%) & .175 & (+203\%) & .259 & (+171\%) & .330 & (+143\%)  \\
      \hline\hline
    \end{tabular}
    }
    }
  \end{center}
\end{table}

The results show that utility-theoretic methods bring about a much
slighter improvement with respect to the baseline, compared to what we
have seen for the macro-oriented methods. For instance, for the SVM
learner, \textsc{Reuters-21578} dataset, and validation depth
$\xi=.10$, the improvement of our (static) micro-oriented
utility-theoretic method with respect to the baseline is just +2\%,
while the improvement was +51\% for the equivalent macro-oriented
method. Across the two ranking methods (static and dynamic), five
datasets, two learners, and three values of inspection depth studied,
improvements range from -1\% (i.e., in a few peculiar cases we even
have a small deterioration) to +14\%, much smaller than in the
macro-oriented case in which the improvements ranged between +2\% and
+402\%.

The main reason for these much smaller improvements lies in the
combined action of two factors. The first factor is that the
validation gains of Equations \ref{eq:gainmicro} are computed on the
global contingency table, whose cells contain very large numbers,
$|\mathcal{C}|$ times larger than the values in the local contingency
tables of the macro-oriented method. This means that, since the values
of the validation gains are very small (given that an increase or a
decrease by 1 of very large values brings about little difference),
the difference between $G(d_{i},fp_{j})$ and $G(d_{i},fn_{j})$ is even
smaller. This makes the difference between the utility-theoretic
methods and the baseline smaller. The second factor is that the
utility function of Equation \ref{eq:utilitymicro}, by collapsing all
the class-specific utility values for a document into a single value,
tends to dwarf the differences between the documents.

It should also be noted that, in the micro-oriented method,
improvements are small also \emph{because the margins of improvement
are small}. To witness, the improvements brought about by
\textsf{Oracle2(d)} (our theoretical upper bound) with respect to the
baseline are smaller than for the macro-oriented method. For instance,
for the \mpb\ learner, \textsc{Reuters-21578} dataset, and validation
depth $\xi=.10$, this improvement is +168\%, while it was +571\% for
the macro-oriented method. So, improving over the baseline is more
difficult for the micro-oriented method than for the macro-oriented
one. The reason why the margins of improvement are smaller is that,
when accuracy is evaluated at the macro level, the infrequent classes
play a bigger role than when evaluating at the micro level. Infrequent
classes are such that a large reduction in error can be achieved even
by validating a few documents of the right type (i.e., false
negatives). As a consequence, for the infrequent classes a ranking
method that pays attention to validation gains has the potential to
obtain sizeable improvements in accuracy right from the beginning; and
a method that favours the infrequent classes tends to shine when
evaluated at the macro level.

\section{Conclusions}
\label{sec:conclusions}

\noindent We have presented a range of methods, all based on utility
theory, for ranking the documents labelled by an automatic
classifier. The documents are ranked in such a way as to maximize the
expected reduction in classification error brought about by a human
annotator who validates a top-ranked subset of the ranked list. We
have also proposed an evaluation measure for such ranking methods,
based on the expectation of the (normalized) reduction in error
brought about by the human annotator's validation activity. This
``semi-automated document classification'' task is different from
``soft (document-ranking) classification'', since in the latter case
it is the documents with the highest probability of being members of
the class (and not the ones which bring about the highest expected
utility if validated) that are top-ranked.


Experiments carried out on standard datasets and variants thereof show
that the intuition of using utility theory is correct. In particular,
of four methods studied, we have found that two methods optimized for
micro-averaged effectiveness bring about only limited improvements,
while the two methods optimized for macro-averaged effectiveness
deliver drastically improved performance with respect to the
baseline. We have also found that the two ``static'' methods, while
seemingly inferior to the ``dynamic'' ones on a purely intuitive
basis, perform as well as the dynamic ones at a fraction of the
computational cost.



It should be remarked that the very fact of using a utility function,
i.e., a function in which different events are characterized by
different gains, makes sense here since we have adopted an evaluation
function, such as $F_{1}$, in which correcting a false positive or a
false negative brings about different benefits to the final
effectiveness score. If we instead adopted standard \emph{accuracy}
(i.e., the percentage of binary classification decisions that are
correct) as the evaluation measure, utility would default to the
probability of misclassification, and our method would coincide with
the baseline, since correcting a false positive or a false negative
would bring about the same benefit. The methods we have presented are
justified by the fact that, in text classification and in other
classification contexts in which imbalance is the rule, $F_{1}$ is the
standard evaluation function, while standard accuracy is a deprecated
measure because of its lack of robustness to class imbalance (see
e.g., \cite[Section 7.1.2]{ACMCS02} for a discussion of this point).

The methods we have proposed are valid also when a different
instantiation of the $F_{\beta}$ function (i.e., with $\beta\not=1$)
is used as the evaluation function. This may be the case, e.g., when
classification is to be applied to a recall-oriented task (such as
e-discovery \cite{Oard:2010fk,Oard:2013fk}), in which case values
$\beta>1$ are appropriate. In these cases our utility-theoretic method
can be used once the appropriate instance of $F_{\beta}$ is plugged,
in place of $F_{1}$, into the equations defining the validation gains
(and into the equations that lead to the definition of
$ENER_{\rho}(\xi)$). The same trivially holds for any other evaluation
function, even different from $F_{\beta}$ and even multivariate and
non-linear, provided it can be computed from a contingency table. It
is easy to foresee that, the higher the difference between the roles
that false positives and false negatives play into the chosen
function, the bigger the improvements brought about by the
utility-theoretic methods with respect to the baseline are going to
be. (For instance, it is easy to foresee that these improvements would
be higher for $F_{2}$ than for $F_{1}$.)

We also remark that this technique is not limited to \emph{text}
classification, but can be useful in any classification context in
which class imbalance \cite{He:2009uq}, or cost-sensitivity in general
\cite{Elkan:2001fk}, suggest using a measure (such as $F_{\beta}$)
that caters for these characteristics.

Note that, by using our methods, it is also easy to provide the human
annotator with an estimate of how accurate the labels of the test set
are as a result of her validation activity. In fact, if the
contingency cell maximum-likelihood estimates $\hat{TP}_{j}^{ML}$,
$\hat{FP}_{j}^{ML}$, and $\hat{FN}_{j}^{ML}$ (see Section
\ref{sec:smoothing}) are updated (adding and subtracting 1 where
appropriate) after each correction by the human annotator,
at any point in the validation activity these are up-to-date estimates
of how well the test set is now classified, and from these estimates
$F_{1}$ (or other) can be computed as usual.

\blue{In the future, we would like to try applying a SATC method after
a transductive learner (e.g., Transductive SVMs
\cite{Joachims:1999fr}) has been used to generate the base classifier
in place of the standard inductive learners we have used in this
work. A transductive method, rather than attempting to generate a
model that minimizes the expected risk on \emph{any} test set,
attempts to minimize misclassifications on a \emph{specific} test
set. When the focus of one's application is squeezing the highest
possible accuracy from a specific test set, as is the case when using
SATC, it would thus make sense to use a transductive instead of an
inductive learning method.}

\section{Acknowledgments} We would like to thank David Lewis and Diego Marcheggiani for many
interesting discussions on the topics of this paper.


\bibliographystyle{ACM-Reference-Format-Journals}
\bibliography{RecentStuff}


\begin{thebibliography}{00}


\ifx \showCODEN    \undefined \def \showCODEN     #1{\unskip}     \fi
\ifx \showDOI      \undefined \def \showDOI       #1{{\tt DOI:}\penalty0{#1}\ }
  \fi
\ifx \showISBNx    \undefined \def \showISBNx     #1{\unskip}     \fi
\ifx \showISBNxiii \undefined \def \showISBNxiii  #1{\unskip}     \fi
\ifx \showISSN     \undefined \def \showISSN      #1{\unskip}     \fi
\ifx \showLCCN     \undefined \def \showLCCN      #1{\unskip}     \fi
\ifx \shownote     \undefined \def \shownote      #1{#1}          \fi
\ifx \showarticletitle \undefined \def \showarticletitle #1{#1}   \fi
\ifx \showURL      \undefined \def \showURL       #1{#1}          \fi

\bibitem[\protect\citeauthoryear{Aalbersberg}{Aalbersberg}{1992}]%
        {Aalbersberg:1992fk}
{IJsbrand~J. Aalbersberg}. 1992.
\newblock \showarticletitle{Incremental Relevance Feedback}. In {\em
  Proceedings of the 15th ACM International Conference on Research and
  Development in Information Retrieval (SIGIR 1992)}. Copenhagen, {DK}, 11--22.
\newblock


\bibitem[\protect\citeauthoryear{Anand}{Anand}{1993}]%
        {Anand:1993fk}
{Paul Anand}. 1993.
\newblock {\em Foundations of Rational Choice under Risk}.
\newblock Oxford University Press, Oxford, UK.
\newblock


\bibitem[\protect\citeauthoryear{Berardi, Esuli, and Sebastiani}{Berardi
  et~al\mbox{.}}{2012}]%
        {Berardi:2012fk}
{Giacomo Berardi}, {Andrea Esuli}, {and} {Fabrizio Sebastiani}. 2012.
\newblock \showarticletitle{A Utility-Theoretic Ranking Method for
  Semi-Automated Text Classification}. In {\em Proceedings of the 35th Annual
  International ACM SIGIR Conference on Research and Development in Information
  Retrieval (SIGIR 2012)}. Portland, {US}, 961--970.
\newblock


\bibitem[\protect\citeauthoryear{Berardi, Esuli, and Sebastiani}{Berardi
  et~al\mbox{.}}{2014}]%
        {Berardi:2014ys}
{Giacomo Berardi}, {Andrea Esuli}, {and} {Fabrizio Sebastiani}. 2014.
\newblock \showarticletitle{Optimising human inspection work in automated
  verbatim coding}.
\newblock {\em International Journal of Market Research\/} {56}, 4 (2014),
  489--512.
\newblock


\bibitem[\protect\citeauthoryear{Brandt, Joachims, Yue, and Bank}{Brandt
  et~al\mbox{.}}{2011}]%
        {Brandt:2011fk}
{Christina Brandt}, {Thorsten Joachims}, {Yisong Yue}, {and} {Jacob Bank}.
  2011.
\newblock \showarticletitle{Dynamic Ranked Retrieval}. In {\em Proceedings of
  the 4th International Conference on Web Search and Web Data Mining (WSDM
  2011)}. Hong Kong, {CN}, 247--256.
\newblock


\bibitem[\protect\citeauthoryear{Brodley and Friedl}{Brodley and
  Friedl}{1999}]%
        {Brodley:1999kx}
{Carla~E. Brodley} {and} {Mark~A. Friedl}. 1999.
\newblock \showarticletitle{Identifying mislabeled training data}.
\newblock {\em Journal of Artificial Intelligence Research\/}  {11} (1999),
  131--167.
\newblock


\bibitem[\protect\citeauthoryear{Burman}{Burman}{1987}]%
        {Burman:1987fk}
{Prabir Burman}. 1987.
\newblock \showarticletitle{Smoothing Sparse Contingency Tables}.
\newblock {\em The Indian Journal of Statistics\/} {49}, 1 (1987), 24--36.
\newblock


\bibitem[\protect\citeauthoryear{Chapelle, Sch{\"o}lkopf, and Zien}{Chapelle
  et~al\mbox{.}}{2006}]%
        {ChaSchZie06}
{Olivier Chapelle}, {Bernard Sch{\"o}lkopf}, {and} {Alexander Zien} (Eds.).
  2006.
\newblock {\em Semi-Supervised Learning}.
\newblock The MIT Press, Cambridge, {US}.
\newblock


\bibitem[\protect\citeauthoryear{Chen and Goodman}{Chen and Goodman}{1996}]%
        {Chen:1996fk}
{Stanley~F. Chen} {and} {Joshua Goodman}. 1996.
\newblock \showarticletitle{An Empirical Study of Smoothing Techniques for
  Language Modeling}. In {\em Proceedings of the 34th Annual Meeting on
  Association for Computational Linguistics (ACL 1996)}. Santa Cruz, {US},
  310--318.
\newblock


\bibitem[\protect\citeauthoryear{Elkan}{Elkan}{2001}]%
        {Elkan:2001fk}
{Charles Elkan}. 2001.
\newblock \showarticletitle{The foundations of cost-sensitive learning}. In
  {\em Proceedings of the 17th International Joint Conference on Artificial
  Intelligence (IJCAI 2001)}. Seattle, {US}, 973--978.
\newblock


\bibitem[\protect\citeauthoryear{Esuli, Fagni, and Sebastiani}{Esuli
  et~al\mbox{.}}{2006}]%
        {Esuli:2006fj}
{Andrea Esuli}, {Tiziano Fagni}, {and} {Fabrizio Sebastiani}. 2006.
\newblock \showarticletitle{{MP-Boost: A} Multiple-Pivot Boosting Algorithm and
  its Application to Text Categorization}. In {\em Proceedings of the 13th
  International Symposium on String Processing and Information Retrieval (SPIRE
  2006)}. Glasgow, UK, 1--12.
\newblock


\bibitem[\protect\citeauthoryear{Esuli and Sebastiani}{Esuli and
  Sebastiani}{2009}]%
        {Esuli:2009th}
{Andrea Esuli} {and} {Fabrizio Sebastiani}. 2009.
\newblock \showarticletitle{Active Learning Strategies for Multi-Label Text
  Classification}. In {\em Proceedings of the 31st European Conference on
  Information Retrieval (ECIR 2009)}. Toulouse, {FR}, 102--113.
\newblock


\bibitem[\protect\citeauthoryear{Esuli and Sebastiani}{Esuli and
  Sebastiani}{2013}]%
        {Esuli:2013ko}
{Andrea Esuli} {and} {Fabrizio Sebastiani}. 2013.
\newblock \showarticletitle{Training Data Cleaning for Text Classification}.
\newblock {\em {ACM} Transactions on Information Systems\/} {31}, 4 (2013).
\newblock


\bibitem[\protect\citeauthoryear{Fukumoto and Suzuki}{Fukumoto and
  Suzuki}{2004}]%
        {Fukumoto:2004gf}
{Fumiyo Fukumoto} {and} {Yoshimi Suzuki}. 2004.
\newblock \showarticletitle{Correcting category errors in text classification}.
  In {\em Proceedings of the 20th International Conference on Computational
  Linguistics (COLING 2004)}. Geneva, {CH}, 868--874.
\newblock


\bibitem[\protect\citeauthoryear{Gale and Church}{Gale and Church}{1994}]%
        {Gale:1994fk}
{William~A. Gale} {and} {Kenneth~W. Church}. 1994.
\newblock \showarticletitle{What's Wrong with Adding One?}
\newblock In {\em Corpus-Based Research into Language: {In} honour of {Jan
  Aarts}}, {N.~Oostdijk} {and} {P.~de~Haan} (Eds.). Rodopi, Amsterdam, {NL},
  189--200.
\newblock


\bibitem[\protect\citeauthoryear{Godbole, Harpale, Sarawagi, and
  Chakrabarti}{Godbole et~al\mbox{.}}{2004}]%
        {Godbole:2004fk}
{Shantanu Godbole}, {Abhay Harpale}, {Sunita Sarawagi}, {and} {Soumen
  Chakrabarti}. 2004.
\newblock \showarticletitle{Document Classification Through Interactive
  Supervision of Document and Term Labels}. In {\em Proceedings of the 8th
  European Conference on Principles and Practice of Knowledge Discovery in
  Databases (PKDD 2004)}. Pisa, {IT}, 185--196.
\newblock


\bibitem[\protect\citeauthoryear{He and Garcia}{He and Garcia}{2009}]%
        {He:2009uq}
{Haibo He} {and} {Edwardo~A. Garcia}. 2009.
\newblock \showarticletitle{Learning from imbalanced data}.
\newblock {\em IEEE Transactions on Knowledge and Data Engineering\/} {21}, 9
  (2009), 1263--1284.
\newblock


\bibitem[\protect\citeauthoryear{Hersh, Buckley, Leone, and Hickman}{Hersh
  et~al\mbox{.}}{1994}]%
        {Hersh94}
{William Hersh}, {Christopher Buckley}, {T.J. Leone}, {and} {David Hickman}.
  1994.
\newblock \showarticletitle{{OHSUMED}: {A}n interactive retrieval evaluation
  and new large text collection for research}. In {\em Proceedings of the 17th
  ACM International Conference on Research and Development in Information
  Retrieval (SIGIR 1994)}. Dublin, {IE}, 192--201.
\newblock


\bibitem[\protect\citeauthoryear{Hoi, Jin, and Lyu}{Hoi et~al\mbox{.}}{2006}]%
        {Hoi:2006ef}
{Steven~C. Hoi}, {Rong Jin}, {and} {Michael~R. Lyu}. 2006.
\newblock \showarticletitle{Large-scale text categorization by batch mode
  active learning}. In {\em Proceedings of the 15th International Conference on
  World Wide Web (WWW 2006)}. Edinburgh, {UK}, 633--642.
\newblock


\bibitem[\protect\citeauthoryear{Ittner, Lewis, and Ahn}{Ittner
  et~al\mbox{.}}{1995}]%
        {Ittner95}
{David~J. Ittner}, {David~D. Lewis}, {and} {David~D. Ahn}. 1995.
\newblock \showarticletitle{Text categorization of low quality images}. In {\em
  Proceedings of the 4th Annual Symposium on Document Analysis and Information
  Retrieval (SDAIR 1995)}. Las Vegas, US, 301--315.
\newblock


\bibitem[\protect\citeauthoryear{Joachims}{Joachims}{1999}]%
        {Joachims:1999fr}
{Thorsten Joachims}. 1999.
\newblock \showarticletitle{Transductive Inference for Text Classification
  using Support Vector Machines}. In {\em Proceedings of the 16th International
  Conference on Machine Learning (ICML 1999)}. Bled, {SL}, 200--209.
\newblock


\bibitem[\protect\citeauthoryear{Kapoor, Horvitz, and Basu}{Kapoor
  et~al\mbox{.}}{2007}]%
        {Kapoor:2007}
{Ashish Kapoor}, {Eric Horvitz}, {and} {Sumit Basu}. 2007.
\newblock \showarticletitle{Selective Supervision: {G}uiding Supervised
  Learning with Decision-Theoretic Active Learning}. In {\em Proceedings of the
  20th International Joint Conference on Artifical Intelligence (IJCAI 2007)}.
  San Francisco, {US}, 877--882.
\newblock


\bibitem[\protect\citeauthoryear{Larkey and Croft}{Larkey and Croft}{1996}]%
        {Larkey96}
{Leah~S. Larkey} {and} {W.~Bruce Croft}. 1996.
\newblock \showarticletitle{Combining classifiers in text categorization}. In
  {\em Proceedings of the 19th ACM International Conference on Research and
  Development in Information Retrieval (SIGIR 1996)}. Z{\"{u}}rich, CH,
  289--297.
\newblock


\bibitem[\protect\citeauthoryear{Lewis and Catlett}{Lewis and Catlett}{1994}]%
        {Lewis94c}
{David~D. Lewis} {and} {Jason Catlett}. 1994.
\newblock \showarticletitle{Heterogeneous uncertainty sampling for supervised
  learning}. In {\em Proceedings of 11th International Conference on Machine
  Learning (ICML 1994)}. New Brunswick, {US}, 148--156.
\newblock


\bibitem[\protect\citeauthoryear{Lewis, Schapire, Callan, and Papka}{Lewis
  et~al\mbox{.}}{1996}]%
        {Lewis96}
{David~D. Lewis}, {Robert~E. Schapire}, {James~P. Callan}, {and} {Ron Papka}.
  1996.
\newblock \showarticletitle{Training algorithms for linear text classifiers}.
  In {\em Proceedings of the 19th ACM International Conference on Research and
  Development in Information Retrieval (SIGIR 1996)}. Z{\"{u}}rich, {CH},
  298--306.
\newblock


\bibitem[\protect\citeauthoryear{Martinez-Alvarez, Bellogin, and
  Roelleke}{Martinez-Alvarez et~al\mbox{.}}{2013}]%
        {Martinez-Alvarez:2013fk}
{Miguel Martinez-Alvarez}, {Alejandro Bellogin}, {and} {Thomas Roelleke}. 2013.
\newblock \showarticletitle{Document Difficulty Framework for Semi-Automatic
  Text Classification}. In {\em Proceedings of the 15th International
  Conference on Data Warehousing and Knowledge Discovery (DaWaK 2013)}. Prague,
  {CZ}.
\newblock


\bibitem[\protect\citeauthoryear{Martinez-Alvarez, Yahyaei, and
  Roelleke}{Martinez-Alvarez et~al\mbox{.}}{2012}]%
        {Martinez-Alvarez:2012fk}
{Miguel Martinez-Alvarez}, {Sirvan Yahyaei}, {and} {Thomas Roelleke}. 2012.
\newblock \showarticletitle{Semi-automatic Document classification:
  {Exploiting} Document Difficulty}. In {\em Proceedings of the 34th European
  Conference on Information Retrieval (ECIR 2012)}. Barcelona, {ES}.
\newblock


\bibitem[\protect\citeauthoryear{McCallum and Nigam}{McCallum and
  Nigam}{1998}]%
        {McCallum98}
{Andrew~K. McCallum} {and} {Kamal Nigam}. 1998.
\newblock \showarticletitle{Employing {EM} in pool-based active learning for
  text classification}. In {\em Proceedings of the 15th International
  Conference on Machine Learning (ICML 1998)}. Madison, {US}, 350--358.
\newblock


\bibitem[\protect\citeauthoryear{Moffat and Zobel}{Moffat and Zobel}{2008}]%
        {Moffat:2008fk}
{Alistair Moffat} {and} {Justin Zobel}. 2008.
\newblock \showarticletitle{Rank-Biased Precision for Measurement of Retrieval
  Effectiveness}.
\newblock {\em {ACM} Transactions on Information Systems\/} {27}, 1 (2008).
\newblock


\bibitem[\protect\citeauthoryear{Niculescu-Mizil and Caruana}{Niculescu-Mizil
  and Caruana}{2005}]%
        {Niculescu-Mizil:2005kx}
{Alexandru Niculescu-Mizil} {and} {Rich Caruana}. 2005.
\newblock \showarticletitle{Obtaining Calibrated Probabilities from Boosting}.
  In {\em Proceedings of the 21st Conference Annual Conference on Uncertainty
  in Artificial Intelligence (UAI 2005)}. Arlington, {US}, 413--420.
\newblock


\bibitem[\protect\citeauthoryear{Oard, Baron, Hedin, Lewis, and Tomlinson}{Oard
  et~al\mbox{.}}{2010}]%
        {Oard:2010fk}
{Douglas~W. Oard}, {Jason~R. Baron}, {Bruce Hedin}, {David~D. Lewis}, {and}
  {Stephen Tomlinson}. 2010.
\newblock \showarticletitle{Evaluation of information retrieval for
  {E-discovery}}.
\newblock {\em Artificial Intelligence and Law\/} {18}, 4 (2010), 347--386.
\newblock


\bibitem[\protect\citeauthoryear{Oard and Webber}{Oard and Webber}{2013}]%
        {Oard:2013fk}
{Douglas~W. Oard} {and} {William Webber}. 2013.
\newblock \showarticletitle{Information Retrieval for E-Discovery}.
\newblock {\em Foundations and Trends in Information Retrieval\/} {7}, 2/3
  (2013).
\newblock


\bibitem[\protect\citeauthoryear{Platt}{Platt}{2000}]%
        {Platt:2000fk}
{John~C. Platt}. 2000.
\newblock \showarticletitle{Probabilistic outputs for support vector machines
  and comparison to regularized likelihood methods}.
\newblock In {\em Advances in Large Margin Classifiers}, {Alexander Smola},
  {Peter Bartlett}, {Bernard Sch{\"o}lkopf}, {and} {Dale Schuurmans} (Eds.).
  The MIT Press, Cambridge, MA, 61--74.
\newblock


\bibitem[\protect\citeauthoryear{Raghavan, Madani, and Jones}{Raghavan
  et~al\mbox{.}}{2006}]%
        {Raghavan:2006:ALF}
{Hema Raghavan}, {Omid Madani}, {and} {Rosie Jones}. 2006.
\newblock \showarticletitle{Active Learning with Feedback on Features and
  Instances}.
\newblock {\em Journal of Machine Learning Research\/}  {7} (2006), 1655--1686.
\newblock


\bibitem[\protect\citeauthoryear{Robertson}{Robertson}{2008}]%
        {Robertson:2008kx}
{Stephen~E. Robertson}. 2008.
\newblock \showarticletitle{A new interpretation of average precision}. In {\em
  Proceedings of the 31st ACM International Conference on Research and
  Development in Information Retrieval (SIGIR 2008)}. Singapore, {SN},
  689--690.
\newblock


\bibitem[\protect\citeauthoryear{Schapire and Singer}{Schapire and
  Singer}{2000}]%
        {Schapire00}
{Robert~E. Schapire} {and} {Yoram Singer}. 2000.
\newblock \showarticletitle{BoosTexter: {A} boosting-based system for text
  categorization}.
\newblock {\em Machine Learning\/} {39}, 2/3 (2000), 135--168.
\newblock


\bibitem[\protect\citeauthoryear{Sebastiani}{Sebastiani}{2002}]%
        {ACMCS02}
{Fabrizio Sebastiani}. 2002.
\newblock \showarticletitle{Machine learning in automated text categorization}.
\newblock {\it Comput. Surveys} {34}, 1 (2002), 1--47.
\newblock


\bibitem[\protect\citeauthoryear{Settles}{Settles}{2012}]%
        {Settles:2012uq}
{Burr Settles}. 2012.
\newblock {\em Active learning}.
\newblock Morgan \& Claypool Publishers, San Rafael, {US}.
\newblock


\bibitem[\protect\citeauthoryear{Simonoff}{Simonoff}{1983}]%
        {Simonoff:1983uq}
{Jeffrey~S. Simonoff}. 1983.
\newblock \showarticletitle{A penalty function approach to smoothing large
  sparse contingency tables}.
\newblock {\em The Annals of Statistics\/} {11}, 1 (1983), 208--218.
\newblock


\bibitem[\protect\citeauthoryear{Tong and Koller}{Tong and Koller}{2001}]%
        {Tong01}
{Simon Tong} {and} {Daphne Koller}. 2001.
\newblock \showarticletitle{Support Vector Machine Active Learning with
  Applications to Text Classification}.
\newblock {\em Journal of Machine Learning Research\/}  {2} (2001), 45--66.
\newblock


\bibitem[\protect\citeauthoryear{Vijayanarasimhan and Grauman}{Vijayanarasimhan
  and Grauman}{2009}]%
        {Vijaya:2009}
{Sudheendra Vijayanarasimhan} {and} {Kristen Grauman}. 2009.
\newblock \showarticletitle{What's it going to cost you?: Predicting effort
  vs.\ informativeness for multi-label image annotations}. In {\em Proceedings
  of the 15th IEEE Conference on Computer Vision and Pattern Recognition (CVPR
  2009)}. Miami, {US}, 2262--2269.
\newblock


\bibitem[\protect\citeauthoryear{{von Neumann} and Morgenstern}{{von Neumann}
  and Morgenstern}{1944}]%
        {von-Neumann:1944uq}
{John {von Neumann}} {and} {Oskar Morgenstern}. 1944.
\newblock {\em Theory of Games and Economic Behavior}.
\newblock Princeton University Press, Princeton, US.
\newblock


\bibitem[\protect\citeauthoryear{Yang and Liu}{Yang and Liu}{1999}]%
        {Yang99}
{Yiming Yang} {and} {Xin Liu}. 1999.
\newblock \showarticletitle{A re-examination of text categorization methods}.
  In {\em Proceedings of the 22nd ACM International Conference on Research and
  Development in Information Retrieval (SIGIR 1999)}. Berkeley, {US}, 42--49.
\newblock


\bibitem[\protect\citeauthoryear{Zhai and Lafferty}{Zhai and Lafferty}{2004}]%
        {Zhai:2004fk}
{ChengXiang Zhai} {and} {John Lafferty}. 2004.
\newblock \showarticletitle{A Study of Smoothing Methods for Language Models
  Applied to Information Retrieval}.
\newblock {\em ACM Transactions on Information Systems\/} {22}, 2 (2004),
  179--214.
\newblock


\bibitem[\protect\citeauthoryear{Zhu and Goldberg}{Zhu and Goldberg}{2009}]%
        {Zhu:2009fk}
{Xiaojin Zhu} {and} {Andrew~B. Goldberg}. 2009.
\newblock {\em Introduction to Semi-Supervised Learning}.
\newblock Morgan and Claypool, San Rafael, US.
\newblock


\end{thebibliography}



\label{@lastpg}

\end{document}